\documentclass{article}

\usepackage{PRIMEarxiv}

\usepackage[utf8]{inputenc}
\usepackage[T1]{fontenc}
\usepackage{times}
\usepackage[round]{natbib}
\usepackage{amsmath,amssymb,booktabs,tabularx,array}
\usepackage{graphicx,microtype}
\usepackage{tikz}
\usetikzlibrary{arrows.meta,positioning}
\usepackage{pgfplots}
\pgfplotsset{compat=1.16}
\usepackage{placeins}
\usepackage{enumitem}
\usepackage{fancyhdr}
\usepackage{url}
\usepackage{hyperref}
\hypersetup{colorlinks=true,linkcolor=blue!60!black,citecolor=blue!60!black,urlcolor=blue!60!black}

\newcommand{\method}{\textsc{SAT}}
\newcommand{\Student}{DeepSeek-V4-Flash}
\newcommand{\Teacher}{Claude-Sonnet-5}
\newcommand{\RQ}[1]{\textbf{RQ#1}}

\title{Teaching the Harness: Student-Aware Skill Transfer for LLM Agents}

\author{
  Chang Liu \\
  Colorado School of Mines \\
  Golden, CO, USA \\
  \texttt{liuchang@mines.edu}
  \And
  Bo Wu \\
  Colorado School of Mines \\
  Golden, CO, USA \\
  \texttt{bwu@mines.edu}
}

\begin{document}
\maketitle

\begin{abstract}
Effective teaching requires reasoning about the learner as well as the task, a capacity that Theory of Mind describes as inferring another agent's latent state from its behavior. We ask whether a stronger LLM teaches a fixed student better when it applies this principle. The student is an LLM agent acting through a harness that fixes its tools, context, and action budget, so a procedure that helps one model can exhaust the budget of another. We introduce Student-Aware Teaching (\method{}). The teacher builds a learner model that pairs an empirical profile with hypotheses about the student's limitations, tests these hypotheses with diagnostic probes, and writes reusable skills for the diagnosed failures. A skill enters the library only after it raises success on matched development tasks. With \Teacher{} teaching \Student{}, \method{} reaches 86.46\% task success across four primary runs, leading both a profile-free teacher with the same framework (81.04\%) and the untaught student (75.00\%) in each run. Removing the analytical learner model lowers both prediction and success, and under misleading learner profiles none of the 48 candidate skills passes admission. When \Student{} and Gemini-3.8-Flash exchange libraries, each performs better with its own. Saved libraries keep average gains of up to 11.1 points after the teacher leaves. The benefit depends on a student that can follow instructions and a deployment route that delivers them.
\end{abstract}


\section{Introduction}
Many large language model (LLM) agents now pair a model with a harness~\citep{yao2023react,yang2024sweagent,wang2025openhands,anthropic2026claudecode,openai2026codex}. A harness is the software layer around the model. It supplies tools, runs a multi-step action loop, manages the context, and caps how many actions the model may take. These components make execution far more complicated than answering a single prompt. A model may know how to restock an inventory and still fail because it spends its allowed actions querying items one at a time. A model may know how to rename files and still leave a nested directory untouched. Many agent failures are failures of execution through a specific harness. This shift also complicates how competence is transferred. Knowledge distillation and machine teaching mostly target the model itself, by updating its weights or choosing its training examples~\citep{hinton2015distilling,hsieh2023distilling,zhu2015machine}. Yet smaller models are increasingly deployed inside harnesses while stronger models remain available at higher cost~\citep{belcak2025slm}, and many deployed models cannot be retrained by their users. The model is also not the only component that can learn. The harness can receive knowledge from an external source as reusable procedures, often called skills, which it loads at run time to guide a fixed model's actions. Skills are readable, cheap to store, and easy to replace. Recent work already learns such skills from execution~\citep{zheng2025skillweaver,yang2026skillopt,alzubi2026evoskill,ni2026trace2skill}. These methods mostly learn from the agent's own failures or optimize skills against a task metric, and less attention has gone to what a teacher should know about the particular model and harness it teaches. We study this question with skills that the harness supplies as advisory context, keeping the model weights and harness code fixed.

The hard part is deciding what a skill should say. Advice to verify every intermediate result helps a student that loses track of its state. The same advice exhausts the action budget of a student that already verifies too often. A useful lesson therefore depends on the learner. Theory of Mind describes the capacity to infer the latent state of another agent from its behavior and to use that inference to anticipate what it will do~\citep{premack1978chimpanzee}. Developmental research links this capacity to teaching, since a teacher must reason about what the learner knows and where it will go wrong~\citep{strauss2002teaching,wellman2004theory}. In practice this reasoning forms a teaching loop. The teacher keeps a working estimate of the student's state, updates it from the student's responses, and uses it to choose the next lesson, so the quality of teaching depends on the quality of that estimate. This view fits harness teaching especially well. The limits that matter live in the execution procedure, such as how many updates fit in the action budget, and a teacher can see them only by modeling how this student acts through this harness. We therefore ask: \emph{does a teacher model write better skills when it first models the student it teaches?}

Student-Aware Teaching (\method{}) takes this teaching loop as its design basis. The learner is a student model paired with its harness. The teacher maintains a learner model with two parts: an empirical profile records what the student does, and an analytical learner model holds hypotheses about why it fails. The teacher tests these hypotheses with controlled probes, such as inventories of increasing size, and writes skills for the diagnosed failures. A candidate enters the library only if it raises the student's success on matched development tasks. After teaching, the teacher leaves and the student works with the saved library alone. Our study focuses on three research questions.
\begin{itemize}[leftmargin=*,itemsep=1pt,topsep=2pt]
\item \RQ{1} (teaching quality). Does an explicit model of the learner produce better skills than teaching from the same types of execution feedback alone?
\item \RQ{2} (teaching content). What does a student-aware teacher teach differently, and are its lessons specific to their student?
\item \RQ{3} (realizing the benefit). Does the curriculum keep helping after the teacher leaves, and which delivery routes, students, and harnesses turn it into a gain?
\end{itemize}

We compare \method{} with an agnostic teacher that uses the same skill generation and admission framework and learns from its own failed trajectories, but has no learner model or probes. \method{} leads it in all four primary runs (\RQ{1}). False-profile and library-swap controls provide additional evidence that teaching benefits from learner-specific information (\RQ{1}, \RQ{2}). Saved libraries keep average gains of 5.6 to 11.1 points with the teacher absent. The gain is realized when the delivery route suits the student and the student can execute procedures. Mid-sized open students benefit most, while the smallest cannot (\RQ{3}). Our contributions are as follows.
\begin{itemize}[leftmargin=*,itemsep=1pt,topsep=2pt]
\item We formulate agent teaching as building a validated skill library for a fixed model and harness, separating teaching quality from how skills are delivered after the teacher leaves.
\item We propose \method{}, which puts a Theory-of-Mind view of teaching into practice. Its learner model pairs an empirical profile with analytical hypotheses, probes test those hypotheses, and matched admission tests validate each lesson.
\item We show that learner-conditioned teaching improves a fixed student over profile-free teaching, particularly when failures arise from execution constraints. Controls with misleading profiles and exchanged libraries support the role of learner-specific information, and student capability and skill delivery determine whether the gain is realized.
\end{itemize}

\section{Related Work}
\label{sec:related}

\paragraph{Learner modeling in teaching.} Theory of Mind research argues that a teacher must reason about a learner's knowledge, ignorance, and misconceptions~\citep{strauss2002teaching,wellman2004theory}. Knowledge tracing and machine teaching formalize instruction for a modeled learner~\citep{corbett1994knowledge,zhu2015machine}, and Iterative Machine Teaching and its black-box extension choose examples from the learner's estimated state~\citep{liu2017iterative,liu2018blackbox}. Machine Theory of Mind and false-belief tests for LLMs study whether models can predict other agents~\citep{rabinowitz2018machine,kosinski2024evaluating}. \method{} uses the idea in a functional sense. Its learner is an LLM paired with its harness, so the state to estimate includes execution limits such as how many updates fit in the budget.

\paragraph{Transferring competence to a fixed model.}
Knowledge distillation and its variants train a student on teacher predictions, rationales, or feedback on the student's own samples~\citep{hinton2015distilling,hsieh2023distilling,agarwal2024onpolicy}. These methods update weights, while \method{} writes an external library that can be inspected and edited and that reaches students with inaccessible weights. Prompt optimization refines instructions or pipelines from textual critiques, task metrics, or execution traces~\citep{pryzant2023protegi,yuksekgonul2024textgrad,khattab2023dspy,agrawal2025gepa}. Our agnostic baseline generates skills from traces, verifier outcomes, and development feedback. \method{} adds an explicit learner model that guides diagnostic probes and skill generation.

\paragraph{Agent memory, skills, and harnesses.}
Experience-based methods store lessons from an agent's own runs as verbal feedback, insights, workflows, or playbooks~\citep{shinn2023reflexion,zhao2024expel,wang2024awm,ouyang2026reasoningbank,zhang2025ace}. Skill libraries range from executable code~\citep{wang2023voyager,wang2025asi} to skills that stronger agents pass to weaker ones~\citep{zheng2025skillweaver}. SkillOpt and EvoSkill revise natural-language skills for a frozen agent from execution failures and keep an edit only when held-out validation improves~\citep{yang2026skillopt,alzubi2026evoskill}. Trace2Skill consolidates lessons from many trajectories into declarative skills that transfer across models~\citep{ni2026trace2skill}. Other work shows that the agent-computer interface shapes performance~\citep{yao2023react,yang2024sweagent} and searches agent, workflow, or harness code~\citep{hu2025adas,zhang2024aflow,lee2026metaharness}, including harnesses built by strong models for weaker targets~\citep{qian2026ai4ai}. \method{} asks what an explicit student model and diagnostic probes add to learning skills from failures. Our library exchange tests how much of a learned library is specific to its student. Table~\ref{tab:related} summarizes how \method{} relates to each family.

\begin{table}[t]
\centering\small
\caption{Positioning of \method{} against related families of work.}
\label{tab:related}
\begin{tabularx}{\linewidth}{l>{\raggedright\arraybackslash}X>{\raggedright\arraybackslash}X>{\raggedright\arraybackslash}X}
\toprule
Family & Examples & What changes & Signal that shapes the change \\
\midrule
Machine teaching & Knowledge tracing, IMT & Training examples & Estimated learner state \\
Distillation & KD, Step-by-Step, GKD & Student weights & Teacher outputs \\
Prompt optimization & ProTeGi, DSPy, TextGrad, GEPA & Prompts and pipelines & Task metric and execution feedback \\
Experience memory & Reflexion, ExpeL, AWM, ReasoningBank, ACE & Memories, workflows, playbooks & The agent's own experience \\
Skill optimization & SkillOpt, EvoSkill, Trace2Skill & Natural-language skills for a frozen agent & Trajectories and validation scores \\
Skills and harnesses & Voyager, SkillWeaver, Meta-Harness, AI4AI & Code skills, agents, or harness code & Environment tests, scores, traces \\
\midrule
\method{} (ours) & & Advisory skills for a fixed model and harness & Learner model with probes, admission tests \\
\bottomrule
\end{tabularx}
\end{table}

\section{Student-Aware Teaching}
\label{sec:method}
\begin{figure}[t]
\centering
\includegraphics[width=\linewidth]{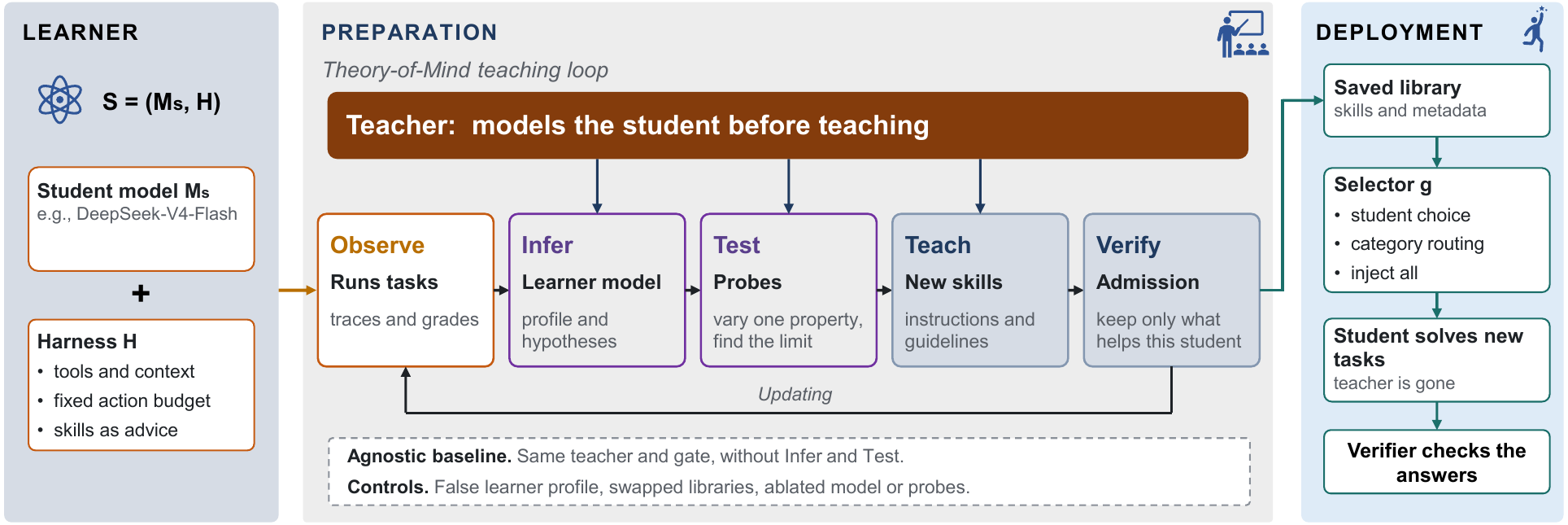}
\caption{Student-Aware Teaching. The learner is a student model operating through its existing harness. During preparation the teacher observes the student, infers a learner model, tests it with diagnostic probes, writes skills for the diagnosed limits, and admits only skills that help the student on matched tasks. At deployment the teacher is absent and a selector delivers the saved skills. The agnostic baseline shares the loop but omits explicit learner modeling and diagnostic probing.}
\label{fig:pipeline}
\end{figure}

\method{} treats teaching as an iterative loop between a stronger teacher and a fixed learner (Figure~\ref{fig:pipeline}). In each round the student attempts teaching tasks, and the teacher updates its model of the student, tests it with diagnostic probes, and writes candidate skills that must pass an admission test. After preparation the teacher leaves, and a selector delivers the saved skills to new tasks.

\subsection{Problem: teaching a fixed learner}
\label{sec:problem}
Let $S=(M_s,H)$ denote a learner formed by a student model $M_s$ and its execution harness $H$. For task $x$ and supplied skills $L_x$, the learner produces a trajectory $\tau\sim S(x,L_x)$, and a verifier returns $V(x,\tau)\in\{0,1\}$. A teaching policy $\pi$ maps preparation evidence $E$ to a skill library $\mathcal L=\pi(E)$. After preparation, the teacher leaves and a selector $g$ chooses which skills accompany each task. Expected performance is
\begin{equation}
J(S,\mathcal L,g)=\mathbb E_{\substack{x\sim\mathcal D\\\tau\sim S(x,g(x,\mathcal L))}}[V(x,\tau)].
\label{eq:objective}
\end{equation}
Teaching aims to raise $J$ under a finite preparation budget, with later instances drawn from task families seen during preparation.

We compare two policies that differ in what the teacher knows about the learner. Both receive the same types of evidence from their own runs: task descriptions, verifier outcomes, execution traces, and development feedback. We write this evidence as $E_{\mathrm{agn}}$ for the agnostic condition and $E$ for the aware condition. The aware policy maintains a learner model $\sigma$ and can request diagnostic probes, so it receives $E_{\mathrm{aware}}=E\cup\{\sigma,\text{probes}\}$ and can predict how this particular student will fail before writing a lesson. Both policies start from the same six generic skills. Our teaching question (\RQ{1}) asks whether $J(S,\pi_{\mathrm{aware}}(E_{\mathrm{aware}}),g)$ exceeds $J(S,\pi_{\mathrm{agn}}(E_{\mathrm{agn}}),g)$ for a fixed learner and selector. Our reuse question (\RQ{3}) fixes the library and varies $g$ with the teacher absent.

\subsection{From student evidence to candidate skills}
\label{sec:evidence}
\paragraph{Learner model.}
The aware teacher maintains a learner model $\sigma$ with two parts. The \emph{empirical profile} records what the student does. At each round the student attempts a batch of teaching tasks with its current library, and the teacher updates category-level success statistics, failure labels, and notes (Appendix~\ref{app:method}). The \emph{analytical learner model} records why the student fails. Observed failures rarely identify their own cause, so the teacher keeps hypotheses about latent execution constraints. Each hypothesis stores a proposed mechanism, a confidence, supporting evidence, and a status. For example, the teacher may hypothesize that repeated stock queries consume the action budget before all updates finish. The learner model also yields a prediction $\hat p_\sigma(x)$ of whether the bare student will solve task $x$. We score these predictions with Brier and AUROC on a fixed calibration panel.

\paragraph{Controlled probes as hypothesis tests.}
Passive observation can support several explanations for one failure. An inventory failure may come from misreading the task, missing a deficient item, or running out of actions. \method{} therefore makes learner modeling active. The teacher requests probes that vary one supported property within a task family, such as inventory size or file count. The bare student attempts each variant, which is rebuilt with its own verifier. If success falls as size grows, the evidence points to an execution-budget limit and the transition records a sampled boundary. The outcome updates the learner model and steers the teacher toward simpler procedures.

\paragraph{Skill generation.}
The teacher proposes up to three candidate skills per round, each a short natural-language procedure with a target category and applicability metadata (Appendix~\ref{app:method}). The aware prompt receives the learner model, weak-category priorities, and probe evidence, while the agnostic prompt receives tasks and failed traces from its own condition. Both forbid copying task-specific answers. Skills are delivered as advice, and the student chooses its own actions.

\subsection{Admission and deployment}
\label{sec:admission}
A candidate $s$ is tested on one to four development tasks $P_s$ from its category. Each task runs once with the candidate alone and once with an empty library. The estimated standalone utility is
\begin{equation}
\widehat U(s)=\frac{1}{|P_s|}\sum_{x\in P_s}\left[Y(x,\{s\})-Y(x,\varnothing)\right],
\label{eq:utility}
\end{equation}
where $Y$ denotes verifier acceptance. A candidate with positive utility enters the library and may replace a generic skill in its category. The library keeps a small number of active skills per category.

After preparation the teacher is removed. We study three selectors $g$. \emph{Student selection} asks the student to choose up to two skills from a metadata menu. \emph{Category routing} matches the task's category tag and picks up to two skills by validation utility. \emph{Flat injection} supplies every active skill. No selector sees test outcomes.

\section{Experimental Setup}
\label{sec:setup}
\paragraph{Base harness.}
Our base harness is a compact planning harness. It first asks the student for a plan and then runs a tool-calling loop of at most ten iterations. It truncates long tool observations and keeps a limited window of recent messages, so action budget and context loss are real constraints. Skills enter as advisory text ahead of the task instruction. Unless a result states another selector, skills reach the student through category routing, both during teaching and in the reported evaluations.

\paragraph{Tasks.}
We use five task suites. A generator builds each task together with its working environment and a programmatic verifier. The suites target different ways an agent can fail inside a tight harness.
\begin{itemize}[leftmargin=*,itemsep=0pt,topsep=2pt]
\item \textbf{Terminal.} Shell tasks such as recursive renaming, sorted numeric merges, and staged file pipelines, which test coverage of the environment.
\item \textbf{Inventory API.} Restocking every item below a threshold through list, query, and restock tools, which tests action economy.
\item \textbf{Arithmetic.} Word problems on discounts, mixtures, and rates, which test task semantics.
\item \textbf{Python code.} Function implementation and bug repair, which test program correctness.
\item \textbf{Dependent workflows.} Chains of file stages that each consume the previous output, which test state tracking.
\end{itemize}

\paragraph{Main setting and experiments.}
Runs S1 through S4 use \Teacher{}~\citep{anthropic2026sonnet5} as teacher and \Student{}~\citep{deepseek2026v4} as student. Each run uses 60 teaching tasks, 30 development tasks, and four teaching rounds. The final evaluation executes 60 held-out tasks, twelve per suite, twice each. Every experiment compares no skill, agnostic teaching, and aware teaching (\method{}) under the same student and harness. \RQ{1} adds two component ablations and a false-profile control, which replaces the learner model with a frozen and deliberately wrong profile. \RQ{2} analyzes skill bodies, an eight-round extended run, and a library swap between \Student{} and Gemini-3.8-Flash~\citep{google2026gemini38}. \RQ{3} evaluates frozen libraries under three selectors with the teacher absent and varies the student across Qwen3~\citep{qwen3}, Qwen2.5-Instruct~\citep{qwen25}, GPT-5.4-Mini~\citep{openai2026gpt54mini}, and Gemini-3.8-Flash. Further experiments cover production harnesses, MATH and GSM8K subsets~\citep{hendrycks2021math,cobbe2021gsm8k}, and self-teaching. Only S1 through S4 enter the primary mean. We report \emph{acceptance}, the percentage of held-out executions whose final result passes the task's verifier (Appendix~\ref{app:details}).

\section{Results}
\label{sec:results}
We present results in the order of the three research questions. Unless noted otherwise, all numbers are final-round acceptance in the base harness with category routing.
\subsection{RQ1: Studying the student makes a better teacher}
\label{sec:rq1}
\begin{table}[t]
\centering\small
\caption{Primary comparison with \Teacher{} teaching \Student{}. Acceptance in percent and gains in points. The top panel reports each run. The bottom panel pools the four runs by task suite, with 48 tasks per suite.}
\label{tab:main}
\begin{tabular}{lrrrrr}
\toprule
 & No skill & Agnostic & Aware & Aware vs.\ none & Aware vs.\ agnostic \\
\midrule
\multicolumn{6}{l}{\emph{Per run}} \\
S1 & 79.17 & 75.83 & \textbf{85.83} & $+6.67$ & $+10.00$ \\
S2 & 77.50 & 85.00 & \textbf{87.50} & $+10.00$ & $+2.50$ \\
S3 & 70.83 & 86.67 & \textbf{87.50} & $+16.67$ & $+0.83$ \\
S4 & 72.50 & 76.67 & \textbf{85.00} & $+12.50$ & $+8.33$ \\
Mean & 75.00 & 81.04 & \textbf{86.46} & $+11.46$ & $+5.42$ \\
Std. & 3.97 & 5.58 & 1.25 & & \\
\midrule
\multicolumn{6}{l}{\emph{Per suite}} \\
Terminal & 53.13 & 62.50 & \textbf{71.88} & $+18.75$ & $+9.38$ \\
Inventory API & 70.83 & 76.04 & \textbf{86.46} & $+15.63$ & $+10.42$ \\
Arithmetic & 80.21 & \textbf{86.46} & 84.38 & $+4.17$ & $-2.08$ \\
Python code & 92.71 & 87.50 & \textbf{94.79} & $+2.08$ & $+7.29$ \\
Dependent workflows & 78.13 & 92.71 & \textbf{94.79} & $+16.67$ & $+2.08$ \\
\bottomrule
\end{tabular}
\end{table}

\textbf{Aware teaching wins every primary run.} Table~\ref{tab:main} reports the primary comparison. \method{} reaches 86.46\% mean acceptance, compared with 81.04\% for agnostic teaching and 75.00\% without skills. Its advantage over agnostic teaching is positive in all four runs and averages 5.42 points, and its gain over no skill averages 11.46 points. Across these four runs, \method{} also has lower observed dispersion, with a standard deviation of 1.25 points against 5.58.

\textbf{The gain concentrates where execution is hard.} The largest aware advantages appear in terminal tasks (71.88\% against 62.50\%) and inventory tasks (86.46\% against 76.04\%), which require the student to cover an environment or complete many updates within a fixed action budget. Arithmetic, where the student mainly needs a formula, slightly favors agnostic teaching. Learner modeling thus helps most when failures come from how the student acts through the harness.

\begin{table}[t]
\centering\small
\caption{Aware acceptance (\%) and the teacher's Brier score for predicting student success on a fixed ten-task panel under different learner information. Each ablation ran as a separate job with its own no-skill and agnostic baselines (Appendix~\ref{app:controls}).}
\label{tab:controls}
\begin{tabular}{lrr}
\toprule
Configuration & Aware acceptance & Teacher Brier \\
\midrule
\multicolumn{3}{l}{\emph{Component ablations, run S2}} \\
Full \method{} & \textbf{87.5} & 0.041 \\
Without fine probes & 83.3 & \textbf{0.036} \\
Without analytical learner model & 83.3 & 0.180 \\
\midrule
\multicolumn{3}{l}{\emph{Misleading profiles, mean of two seeds}} \\
Overestimating profile & 75.0 & 0.238 \\
Underestimating profile & 79.2 & 0.747 \\
\bottomrule
\end{tabular}
\end{table}

\textbf{Component ablations are consistent with benefits from modeling and probing.} Table~\ref{tab:controls} varies what the teacher knows about its student. Final aware acceptance is 83.3\% in both ablated configurations, compared with 87.5\% for full \method{}. Removing the analytical learner model also raises the teacher's Brier score for predicting student success from 0.041 to 0.180. Removing probes leaves prediction accurate, which suggests that prediction accuracy alone does not capture all information relevant to teaching. Probes supply measured limits, such as the inventory size at which the student breaks. Because each configuration ran as its own job with its own baselines, these differences are indicative. Both ablated teachers still exceed their own no-skill and agnostic baselines.

\textbf{Misleading learner profiles derail teaching.} To test sensitivity to inaccurate learner information, we supply deliberately overestimating and underestimating profiles and keep everything else fixed. One profile describes \Student{} as a frontier model for which step limits hardly matter. The other describes it as a tiny model that needs exhaustive scaffolding. Both stay frozen while the teacher observes the real student's trajectories. Under both profiles, none of the 48 generated candidates passes the admission test, so the libraries retain only the six initial generic skills. The teacher's predictions also degrade. The underestimating teacher ends with Brier near 0.75 and AUROC below 0.3, so its predictions run against the student's real outcomes. Final acceptance averages 75.0\% and 79.2\%, within the range of the untaught student in the primary runs. The false-profile teachers spend \$0.36 to \$0.41 per run, similar to the \$0.39 of the correct-profile teacher.

\subsection{RQ2: The aware teacher writes lessons that fit the learner}
\label{sec:rq2}
\textbf{From a diagnosed budget limit to a bulk procedure.} The clearest sequence comes from a Qwen3-8B student taught by Qwen3-14B. After two inventory failures at the action limit, the aware teacher probes inventory size and finds that the student passes with two items and fails at five or more. It then writes a procedure that begins with one full stock query, which passes three of four admission tasks where the bare student passes none. The agnostic plan checks stock item by item (Table~\ref{tab:instructioncase}). Final inventory acceptance is 91.67\% with aware teaching, 0\% with agnostic teaching, and 8.33\% without skills. The same repair appears in S4, where a bulk snapshot skill lifts inventory acceptance to 79.17\% against 50.00\% agnostic. In these cases the learner model changes the procedure itself: per-item or bulk queries, shallow or recursive discovery, and early or deferred checks (Appendix~\ref{app:curriculum}).

\begin{table}[t]
\centering\small
\caption{What a lesson looks like: the retained inventory skills for the Qwen3-8B student, paraphrased from complete skill bodies. Final inventory acceptance is 0\% with the agnostic plan and 91.67\% with the aware skills.}
\label{tab:instructioncase}
\begin{tabularx}{\linewidth}{lX}
\toprule
Curriculum record & Procedure encoded in the skill \\
\midrule
Agnostic plan & List items and inspect each item's stock. Restock deficient items to the threshold. Track processed items and write the completion marker after processing all items. \\
Aware skill 1 & Retrieve all stock in one query. Restock items below the threshold and inspect each updated item. Retain intermediate state for later checks. \\
Aware skill 2 & Retrieve all stock in one query. Restock the required quantities and verify final levels after all updates. \\
\bottomrule
\end{tabularx}
\end{table}

\begin{figure}[t]
\centering
\begin{minipage}[c]{0.47\linewidth}
\centering
\begin{tikzpicture}
\begin{axis}[width=\linewidth,height=4.2cm,xlabel={Teaching round},ylabel={Test acceptance (\%)},
xmin=0.7,xmax=8.3,ymin=73,ymax=88.5,xtick={1,...,8},ytick={74,76,78,80,82,84},
legend columns=2,legend cell align=left,legend style={font=\scriptsize,at={(0.02,0.98)},anchor=north west,draw=none,fill=white,fill opacity=0.85,text opacity=1,/tikz/column 2/.style={column sep=10pt}},
tick label style={font=\scriptsize},label style={font=\small},grid=major,grid style={gray!20}]
\addplot[thick,mark=*,mark size=1.3pt,color=blue!70!black] coordinates {(1,78.3)(2,78.7)(3,80.7)(4,78.7)(5,80.3)(6,80.7)(7,82.3)(8,84.3)};
\addplot[thick,mark=triangle*,mark size=1.5pt,color=gray] coordinates {(1,74.3)(2,75.3)(3,78.0)(4,74.3)(5,76.7)(6,76.0)(7,77.3)(8,78.0)};
\addplot[thick,mark=square*,mark size=1.2pt,color=orange!85!black] coordinates {(1,78.3)(2,79.0)(3,80.0)(4,80.0)(5,79.0)(6,79.0)(7,80.3)(8,80.7)};
\legend{Aware (\method{}),No skill,Agnostic}
\end{axis}
\end{tikzpicture}
\end{minipage}\hfill
\begin{minipage}[c]{0.47\linewidth}
\centering
\begin{tikzpicture}
\begin{axis}[width=\linewidth,height=4.2cm,xlabel={Teaching round},ylabel={Teacher Brier score},
xmin=0.7,xmax=8.3,ymin=0,ymax=0.22,xtick={1,...,8},ytick={0,0.05,0.10,0.15,0.20},
yticklabel style={/pgf/number format/fixed,/pgf/number format/precision=2},
tick label style={font=\scriptsize},label style={font=\small},grid=major,grid style={gray!20}]
\addplot[thick,mark=*,mark size=1.3pt,color=blue!70!black] coordinates {(1,0.132)(2,0.071)(3,0.046)(4,0.064)(5,0.063)(6,0.064)(7,0.079)(8,0.068)};
\addplot[thick,mark=square*,mark size=1.2pt,color=orange!85!black] coordinates {(1,0.175)(2,0.204)(3,0.181)(4,0.128)(5,0.111)(6,0.203)(7,0.108)(8,0.162)};
\end{axis}
\end{tikzpicture}
\end{minipage}
\caption{Eight-round extended teaching with \Teacher{} and \Student{}. Left: test acceptance on 150 held-out tasks after each round. Right: the teacher's Brier score for predicting student success on a fixed ten-task panel (lower is better).}
\label{fig:extended}
\end{figure}
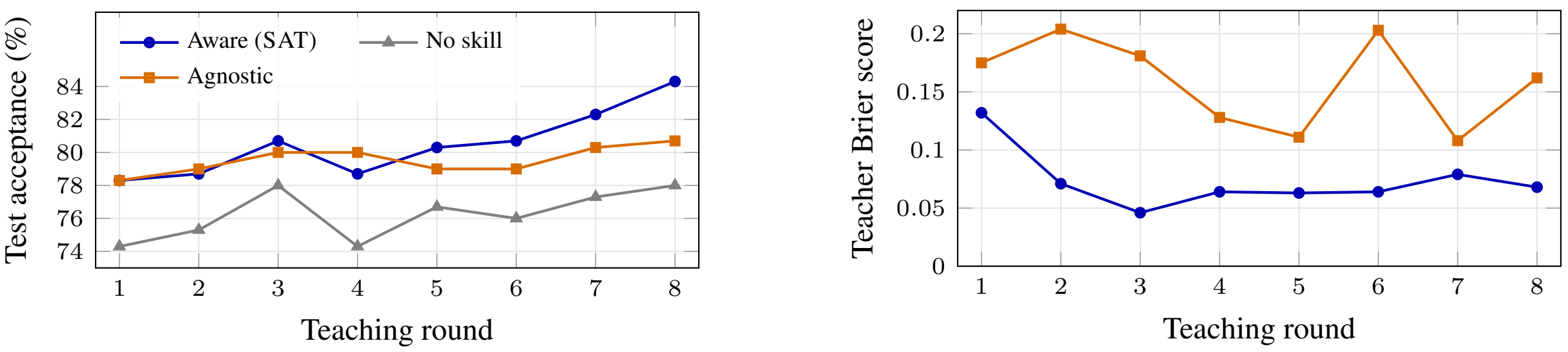

\textbf{From a missed nested file to recursive coverage.} The extended run shows the teacher adapting over eight rounds (Figure~\ref{fig:extended}). In round six, a teaching attempt leaves a nested file unrenamed. The aware teacher responds with a skill that enumerates matching files recursively and checks that no original extension remains. Renaming acceptance rises from 30\% to 100\% and stays there for the final three rounds, while the agnostic student keeps a shallow glob and passes two of ten executions. Overall, aware acceptance climbs from 78.3\% to 84.3\%, against 80.7\% agnostic and 78.0\% without skills. The aware teacher also predicts its student better. Its Brier score is lower in every round, and its predictions separate passing from failing panel tasks perfectly (AUROC 1.0) in six of eight rounds. Its admitted skills have mean standalone utility 0.58 against 0.35 for agnostic skills.

\textbf{Probe results enter the lesson.} The extended run also shows a measured limit passing directly into skill text. In round three, an inventory probe finds that the bare student succeeds with eight items and fails with twelve and sixteen. The skill written in round four cites the twelve-item failure explicitly. It advises one consolidated stock query, progress tracking across updates, and a final check only when the budget allows. The lesson thus builds on an observed failure from the diagnostic probe. A later probe in the same run passes at all tested sizes, so such boundaries depend on the sampled instances.

\textbf{Awareness changes what the curriculum emphasizes.} We annotated the complete main-setting skill bodies with five overlapping functional labels (Appendix~\ref{app:curriculum}). Half of the aware skills teach environment coverage, compared with one in ten agnostic skills. Both curricula give similar amounts of advice on execution economy and verification, yet they differ in when and what to check. Some aware skills defer checks to save actions, while the Qwen3-8B skills add explicit checks during or after stock updates. Verification timing is thus a teaching decision that depends on the learner.

\begin{table}[t]
\centering\small
\caption{Library exchange between two students taught by \Teacher{} in the same harness. Acceptance (\%) with the student's own aware library and with the other student's library, and the paired difference with its 95\% interval. No-skill baselines are 79.2\% for \Student{} and 87.5\% for Gemini-3.8-Flash.}
\label{tab:swap}
\begin{tabular}{llrrl}
\toprule
Student & Route & Own & Exchanged & Own minus exchanged \\
\midrule
\Student{} & Student selection & 79.2 & 75.0 & $+4.2\;[-1.7,10.8]$ \\
 & Category routing & \textbf{93.3} & 83.3 & $+10.0\;[2.5,18.3]$ \\
 & Flat injection & 86.7 & 82.5 & $+4.2\;[-4.2,13.3]$ \\
\addlinespace
Gemini-3.8-Flash & Student selection & 91.7 & 88.3 & $+3.3\;[0.8,6.7]$ \\
 & Category routing & \textbf{93.3} & 90.0 & $+3.3\;[0.0,7.5]$ \\
 & Flat injection & 93.3 & 91.7 & $+1.7\;[0.0,5.0]$ \\
\bottomrule
\end{tabular}
\end{table}

\textbf{Libraries carry student-specific advantages.} If aware lessons encode properties of one learner, they should fit that learner better than another. We exchange the aware libraries of \Student{} and Gemini-3.8-Flash, both taught by the same teacher in the same harness (Table~\ref{tab:swap}). Both students perform better with their own libraries than with the exchanged libraries across the evaluated routes. Category routing fixes retrieval and isolates library content. Under this route \Student{} reaches 93.3\% with its own library and 83.3\% with Gemini's, a difference of 10.0 points with paired interval $[2.5,18.3]$. Several exchanged libraries still help their recipients, and Gemini-3.8-Flash shows smaller margins near its ceiling. The pattern suggests that learner-conditioned teaching produces libraries with some student-specific advantages (Appendix~\ref{app:controls}).

\subsection{RQ3: Realizing the benefit after the teacher leaves}
\label{sec:rq3}
Teaching pays off only if the saved curriculum reaches a student that can use it. We examine three conditions in turn: the absence of the teacher, the route that delivers skills, and the student and harness that receive them.

\begin{table}[t]
\centering\small
\caption{Teacher-free gains from frozen aware libraries, in points over a no-skill baseline run inside each job. L1 to L3 come from the main setting and ``Extended'' is the eight-round library. Local teachers are Qwen2.5-32B-Instruct and Qwen3-14B. Absolute rates and intervals are in Appendix~\ref{app:modes}.}
\label{tab:routes}
\begin{tabular}{llrrr}
\toprule
Student & Library & Student selection & Category routing & Flat injection \\
\midrule
\Student{} & L1 & $+7.5$ & $+9.2$ & $+9.2$ \\
\Student{} & L2 & $-0.8$ & $+7.5$ & $+10.8$ \\
\Student{} & L3 & $+10.0$ & $+9.2$ & $+13.3$ \\
\Student{} & Mean of L1 to L3 & $+5.6$ & $+8.6$ & $+11.1$ \\
\Student{} & Extended & $+5.7$ & $+7.7$ & $+7.0$ \\
\midrule
Qwen2.5-7B-Instruct & Local & $+8.3$ & $+4.2$ & $-14.2$ \\
Qwen3-8B & Local & $0.0$ & $+15.0$ & $+15.0$ \\
\bottomrule
\end{tabular}
\end{table}

\textbf{The curriculum outlives the teacher.} Table~\ref{tab:routes} freezes saved aware libraries and removes the teacher. Every \Student{} library improves the student under category routing and flat injection. Across L1 to L3, mean gains are 5.6 points with student selection, 8.6 with category routing, and 11.1 with flat injection. The library from the extended run keeps gains of 5.7 to 7.7 points. Teaching is thus a one-time investment that keeps paying off without further teacher calls.

\textbf{Delivery can erase a good curriculum.} The local students reveal two failure modes. Qwen3-8B never selects a skill in any of its 120 student-selection executions and gains nothing. Category routing and flat injection deliver the same library and recover a 15-point gain. Qwen2.5-7B-Instruct shows the opposite pattern. It gains 8.3 points when it chooses its own skills and loses 14.2 points when it receives the whole library. The best route therefore depends on the student, and a deployment should choose it with the student in mind.

\begin{table}[t]
\centering\small
\caption{Acceptance (\%) across student and teacher configurations. Qwen2.5 models are Instruct variants. GPT-5.4-Mini is the mean of three runs.}
\label{tab:modelgen}
\begin{tabular}{llrrr}
\toprule
Student & Teacher & No skill & Agnostic & Aware \\
\midrule
Qwen3-8B & Qwen3-14B & 38.3 & 33.3 & \textbf{53.3} \\
Qwen3-8B & Qwen3-32B & 38.3 & 45.0 & \textbf{50.0} \\
Qwen2.5-7B & Qwen2.5-32B & 43.3 & 45.0 & \textbf{46.7} \\
Qwen2.5-3B, run A & Qwen2.5-32B & 16.7 & \textbf{21.7} & 20.0 \\
Qwen2.5-3B, run B & Qwen2.5-32B & 15.0 & 13.3 & 15.0 \\
Qwen3-1.7B & Qwen3-32B & 11.7 & 11.7 & 11.7 \\
GPT-5.4-Mini & \Teacher{} & 88.9 & 90.6 & 90.6 \\
Gemini-3.8-Flash & \Teacher{} & 87.5 & \textbf{92.5} & 91.7 \\
\bottomrule
\end{tabular}
\end{table}

\textbf{Mid-sized students gain the most.} Table~\ref{tab:modelgen} varies the student and teacher. With a Qwen3-14B teacher, Qwen3-8B rises from 38.3\% without skills to 53.3\% with aware teaching, while agnostic teaching falls to 33.3\%. A Qwen3-32B teacher also gives this student a clear aware gain. The smallest students do not respond. Qwen3-1.7B stays at 11.7\% in every condition and retains no learned skill, and Qwen2.5-3B-Instruct changes little across two runs. Strong API students sit at the other end. GPT-5.4-Mini and Gemini-3.8-Flash start near 88\% and gain at most five points from either teacher, and agnostic teaching matches or slightly exceeds aware teaching for both. For GPT-5.4-Mini, the aware teacher predicts its student better in two of three runs while accuracy ties (Appendix~\ref{app:modelstate}). Procedural teaching thus needs a student that can execute procedures and still has room to improve.

\textbf{Portability and exploratory settings.} \method{} can also be applied through the native interfaces of four production harnesses: Codex~\citep{openai2026codex}, Pi~\citep{zechner2026pi}, OpenCode~\citep{opencode2026}, and Claude Code~\citep{anthropic2026claudecode}. Skills are supplied as task context, while each harness retains its own tools and execution workflow. The harnesses differ from the base harness and from each other, so we compare each condition only with the same harness without skills. Final point estimates favor aware teaching over no skill in every harness, by 1.0 to 5.2 points (Appendix~\ref{app:harness}). Appendix~\ref{app:other} further reports exploratory results on small public mathematics subsets and a self-teaching comparison. On MATH and GSM8K subsets, aware teaching exceeds agnostic teaching by 8.3 points on each and matches or exceeds both baselines in all five runs. In the self-teaching comparison, the student's own backbone is already a useful teacher, since self-improvement adds 9.7 points over no skill. Replacing the student's self-generated skills with those from \Teacher{}, while keeping the same inputs, adds a further 1.4 points, which is one additional accepted execution among 72 test executions.

\section{Discussion}
\label{sec:discussion}
\textbf{What the learner model adds.} Both teachers see the student fail, and both can write useful procedures. The learner model changes which procedure the teacher chooses. A failed inventory task is consistent with misreading the task, missing an item, or running out of actions, and each reading suggests a different lesson. When the teacher's model and probes point to the action budget, it writes a bulk query. When they point to incomplete coverage, it writes recursive discovery. The results line up with this account. Gains concentrate in suites where execution constraints bind, ablating the analytical model or the probes lowers acceptance, misleading profiles yield no admissible skill, and each student does better with its own library. In the language of Theory of Mind, the teacher benefits from anticipating how this particular student will act. None of these controls alone isolates learner modeling as the only source of the gain, but they point in the same direction as the main comparison.

\textbf{Scope.} Most evidence comes from one teacher and student pair in one base harness, with few seeds for the controls and deliberately extreme false profiles. Deployment results depend on the selector, and the production and exploratory experiments carry the caveats stated beside them. The learner model also has natural limits: it cannot help a student too small to execute procedures, and it adds little for a student near ceiling. Matched replays of teaching evidence, milder profile errors, more seeds, and dedicated studies in production harnesses are the natural next steps.

\section{Conclusion}
We asked whether a teacher model teaches better when it first models its student. Across four primary runs, Student-Aware Teaching improves over profile-free teaching, with the largest gains on tasks that involve constrained execution. Its saved libraries keep helping after the teacher leaves. Ablations, misleading profiles, and exchanged libraries all support the role of learner-specific information, which Theory of Mind research places at the center of human teaching. Modeling whom it is teaching can help a teacher decide which procedures to teach.
\label{maintext:end}

\section*{AI Use Statement}
We used generative AI as part of the research pipeline described in the paper, including model-based generation and annotation procedures. We also used generative AI tools to polish the manuscript and check its clarity and consistency. All research design, analyses, interpretations, and conclusions were reviewed and verified by the authors, who take full responsibility for the final content.

\section*{Reproducibility Statement}
Section~\ref{sec:method} specifies the teaching and deployment protocol, and Appendix~\ref{app:method} provides implementation details such as prompt limits and admission rules. Appendix~\ref{app:tasks} describes the base harness, task generators, and split sizes. Appendices~\ref{app:modes} and~\ref{app:other} report per-route and per-run results with paired intervals. Appendix~\ref{app:curriculum} gives the annotation codebook and paired skill artifacts. Appendix~\ref{app:extended} reports the full extended trajectory and the paired bootstrap procedure. Appendices~\ref{app:controls} and~\ref{app:harness} describe the learner-specificity controls and the production-harness protocol. Appendix~\ref{app:details} lists the cost ledgers and category-level results.

\clearpage
\bibliographystyle{plainnat}
\bibliography{references}

@inproceedings{agrawal2025gepa,
  title={Gepa: Reflective prompt evolution can outperform reinforcement learning},
  author={Agrawal, Lakshya A and Tan, Shangyin and Soylu, Dilara and Ziems, Noah and Khare, Rishi and Opsahl-Ong, Krista and Singhvi, Arnav and Shandilya, Herumb and Ryan, Michael J and Jiang, Meng and others},
  booktitle={International Conference on Learning Representations},
  volume={2026},
  pages={8479--8565},
  year={2026}
}

@article{hinton2015distilling,
  title={Distilling the knowledge in a neural network},
  author={Hinton, Geoffrey and Vinyals, Oriol and Dean, Jeff},
  journal={arXiv preprint arXiv:1503.02531},
  year={2015}
}

@inproceedings{hsieh2023distilling,
  title={Distilling step-by-step! outperforming larger language models with less training data and smaller model sizes},
  author={Hsieh, Cheng-Yu and Li, Chun-Liang and Yeh, Chih-Kuan and Nakhost, Hootan and Fujii, Yasuhisa and Ratner, Alex and Krishna, Ranjay and Lee, Chen-Yu and Pfister, Tomas},
  booktitle={Findings of the association for computational linguistics: ACL 2023},
  pages={8003--8017},
  year={2023}
}

@inproceedings{hu2025adas,
  title={Automated design of agentic systems},
  author={Hu, Shengran and Lu, Cong and Clune, Jeff},
  booktitle={International Conference on Learning Representations},
  volume={2025},
  pages={21344--21377},
  year={2025}
}

@article{khattab2023dspy,
  title={Dspy: Compiling declarative language model calls into self-improving pipelines},
  author={Khattab, Omar and Singhvi, Arnav and Maheshwari, Paridhi and Zhang, Zhiyuan and Santhanam, Keshav and Vardhamanan, Sri and Haq, Saiful and Sharma, Ashutosh and Joshi, Thomas T and Moazam, Hanna and others},
  journal={arXiv preprint arXiv:2310.03714},
  year={2023}
}

@article{lee2026metaharness,
  title={Meta-harness: End-to-end optimization of model harnesses},
  author={Lee, Yoonho and Nair, Roshen and Zhang, Qizheng and Lee, Kangwook and Khattab, Omar and Finn, Chelsea},
  journal={arXiv preprint arXiv:2603.28052},
  year={2026}
}

@article{qian2026ai4ai,
  title={AI4AI at Test-Time: Strong-to-Weak Capability Transfer via Harnesses},
  author={Qian, Cheng and Zhao, Wenting and Yang, Liangwei and Wang, Heng and Qiu, Jielin and Ji, Heng and Savarese, Silvio and Wang, Huan and Heinecke, Shelby},
  journal={arXiv preprint arXiv:2608.12307},
  year={2026}
}

@article{shinn2023reflexion,
  title={Reflexion: Language agents with verbal reinforcement learning},
  author={Shinn, Noah and Cassano, Federico and Gopinath, Ashwin and Narasimhan, Karthik and Yao, Shunyu},
  journal={Advances in neural information processing systems},
  volume={36},
  pages={8634--8652},
  year={2023}
}

@article{wang2023voyager,
  title={Voyager: An open-ended embodied agent with large language models},
  author={Wang, Guanzhi and Xie, Yuqi and Jiang, Yunfan and Mandlekar, Ajay and Xiao, Chaowei and Zhu, Yuke and Fan, Linxi and Anandkumar, Anima},
  journal={arXiv preprint arXiv:2305.16291},
  year={2023}
}

@inproceedings{zhang2025ace,
  title={Agentic context engineering: Evolving contexts for self-improving language models},
  author={Zhang, Qizheng and Hu, Changran and Upasani, Shubhangi and Ma, Boyuan and Hong, Fenglu and Kamanuru, Vamsidhar and Rainton, Jay and Wu, Chen and Ji, Mengmeng and Li, Hanchen and others},
  booktitle={International Conference on Learning Representations},
  volume={2026},
  pages={86069--86100},
  year={2026}
}

@inproceedings{zhao2024expel,
  title={Expel: Llm agents are experiential learners},
  author={Zhao, Andrew and Huang, Daniel and Xu, Quentin and Lin, Matthieu and Liu, Yong-Jin and Huang, Gao},
  booktitle={Proceedings of the AAAI Conference on Artificial Intelligence},
  volume={38},
  number={17},
  pages={19632--19642},
  year={2024}
}

@article{zheng2025skillweaver,
  title={Skillweaver: Web agents can self-improve by discovering and honing skills},
  author={Zheng, Boyuan and Fatemi, Michael Y and Jin, Xiaolong and Wang, Zora Zhiruo and Gandhi, Apurva and Song, Yueqi and Gu, Yu and Srinivasa, Jayanth and Liu, Gaowen and Neubig, Graham and others},
  journal={arXiv preprint arXiv:2504.07079},
  year={2025}
}

@article{cobbe2021gsm8k,
  title={Training verifiers to solve math word problems},
  author={Cobbe, Karl and Kosaraju, Vineet and Bavarian, Mohammad and Chen, Mark and Jun, Heewoo and Kaiser, Lukasz and Plappert, Matthias and Tworek, Jerry and Hilton, Jacob and Nakano, Reiichiro and others},
  journal={arXiv preprint arXiv:2110.14168},
  year={2021}
}

@article{hendrycks2021math,
  title={Measuring mathematical problem solving with the math dataset},
  author={Hendrycks, Dan and Burns, Collin and Kadavath, Saurav and Arora, Akul and Basart, Steven and Tang, Eric and Song, Dawn and Steinhardt, Jacob},
  journal={arXiv preprint arXiv:2103.03874},
  year={2021}
}

@article{corbett1994knowledge,
  title={Knowledge tracing: Modeling the acquisition of procedural knowledge},
  author={Corbett, Albert T and Anderson, John R},
  journal={User modeling and user-adapted interaction},
  volume={4},
  number={4},
  pages={253--278},
  year={1994},
  publisher={Springer}
}

@inproceedings{zhu2015machine,
  title={Machine teaching: An inverse problem to machine learning and an approach toward optimal education},
  author={Zhu, Xiaojin},
  booktitle={Proceedings of the AAAI conference on artificial intelligence},
  volume={29},
  number={1},
  year={2015}
}

@misc{qwen3,
      title={Qwen3 Technical Report}, 
      author={An Yang and Anfeng Li and Baosong Yang and Beichen Zhang and Binyuan Hui and Bo Zheng and Bowen Yu and Chang Gao and Chengen Huang and Chenxu Lv and others},
      year={2025},
      eprint={2505.09388},
      archivePrefix={arXiv},
      primaryClass={cs.CL},
      url={https://arxiv.org/abs/2505.09388}, 
}

@misc{qwen25,
      title={Qwen2.5 Technical Report}, 
      author={Qwen: An Yang and Baosong Yang and Beichen Zhang and Binyuan Hui and Bo Zheng and Bowen Yu and Chengyuan Li and Dayiheng Liu and Fei Huang and Haoran Wei and others},
      year={2025},
      eprint={2412.15115},
      archivePrefix={arXiv},
      primaryClass={cs.CL},
      url={https://arxiv.org/abs/2412.15115}, 
}

@inproceedings{liu2017iterative,
  title={Iterative machine teaching},
  author={Liu, Weiyang and Dai, Bo and Humayun, Ahmad and Tay, Charlene and Yu, Chen and Smith, Linda B and Rehg, James M and Song, Le},
  booktitle={International Conference on Machine Learning},
  pages={2149--2158},
  year={2017},
  organization={PMLR}
}

@inproceedings{liu2018blackbox,
  title={Towards black-box iterative machine teaching},
  author={Liu, Weiyang and Dai, Bo and Li, Xingguo and Liu, Zhen and Rehg, James and Song, Le},
  booktitle={International Conference on Machine Learning},
  pages={3141--3149},
  year={2018},
  organization={PMLR}
}

@inproceedings{agarwal2024onpolicy,
  title={On-policy distillation of language models: Learning from self-generated mistakes},
  author={Agarwal, Rishabh and Vieillard, Nino and Zhou, Yongchao and Stanczyk, Piotr and Ramos Garea, Sabela and Geist, Matthieu and Bachem, Olivier},
  booktitle={International Conference on Learning Representations},
  volume={2024},
  pages={21246--21263},
  year={2024}
}

@inproceedings{pryzant2023protegi,
  title={Automatic prompt optimization with “gradient descent” and beam search},
  author={Pryzant, Reid and Iter, Dan and Li, Jerry and Lee, Yin and Zhu, Chenguang and Zeng, Michael},
  booktitle={Proceedings of the 2023 conference on empirical methods in natural language processing},
  pages={7957--7968},
  year={2023}
}

@article{yuksekgonul2024textgrad,
  title={Textgrad: Automatic "differentiation" via text},
  author={Yuksekgonul, Mert and Bianchi, Federico and Boen, Joseph and Liu, Sheng and Huang, Zhi and Guestrin, Carlos and Zou, James},
  journal={arXiv preprint arXiv:2406.07496},
  year={2024}
}

@article{wang2024awm,
  title={Agent workflow memory},
  author={Wang, Zora Zhiruo and Mao, Jiayuan and Fried, Daniel and Neubig, Graham},
  journal={arXiv preprint arXiv:2409.07429},
  year={2024}
}

@inproceedings{ouyang2026reasoningbank,
 author = {Ouyang, Siru and Yan, Jun and Hsu, I-Hung and Chen, Yanfei and Jiang, Ke and Wang, Zifeng and Han, Rujun and Le, Long and Daruki, Samira and Tang, Xiangru and others},
 booktitle = {International Conference on Learning Representations},
 editor = {C. Vondrick and B. Hariharan and C. Raffel and L. Pinto and D. Yang and A. Faust},
 pages = {94327--94354},
 title = {ReasoningBank: Scaling Agent Self-Evolving with Reasoning Memory},
 url = {https://proceedings.iclr.cc/paper_files/paper/2026/file/980ea04d23d1f6908964eba2a74afe45-Paper-Conference.pdf},
 volume = {2026},
 year = {2026}
}

@article{wang2025asi,
  title={Inducing programmatic skills for agentic tasks},
  author={Wang, Zora Zhiruo and Gandhi, Apurva and Neubig, Graham and Fried, Daniel},
  journal={arXiv preprint arXiv:2504.06821},
  year={2025}
}

@article{yao2023react,
  title={React: Synergizing reasoning and acting in language models},
  author={Yao, Shunyu and Zhao, Jeffrey and Yu, Dian and Du, Nan and Shafran, Izhak and Narasimhan, Karthik and Cao, Yuan},
  journal={arXiv preprint arXiv:2210.03629},
  year={2022}
}

@article{yang2024sweagent,
  title={Swe-agent: Agent-computer interfaces enable automated software engineering},
  author={Yang, John and Jimenez, Carlos and Wettig, Alexander and Lieret, Kilian and Yao, Shunyu and Narasimhan, Karthik and Press, Ofir},
  journal={Advances in Neural Information Processing Systems},
  volume={37},
  pages={50528--50652},
  year={2024}
}

@inproceedings{zhang2024aflow,
  title={Aflow: Automating agentic workflow generation},
  author={Zhang, Jiayi and Xiang, Jinyu and Yu, Zhaoyang and Teng, Fengwei and Chen, Xionghui and Chen, Jiaqi and Zhuge, Mingchen and Cheng, Xin and Hong, Sirui and Wang, Jinlin and others},
  booktitle={International Conference on Learning Representations},
  volume={2025},
  pages={34040--34077},
  year={2025}
}

@article{premack1978chimpanzee,
  title={Does the chimpanzee have a theory of mind?},
  author={Premack, David and Woodruff, Guy},
  journal={Behavioral and brain sciences},
  volume={1},
  number={4},
  pages={515--526},
  year={1978},
  publisher={Cambridge University Press}
}

@article{strauss2002teaching,
  title={Teaching as a natural cognition and its relations to preschoolers’ developing theory of mind},
  author={Strauss, Sidney and Ziv, Margalit and Stein, Adi},
  journal={Cognitive development},
  volume={17},
  number={3-4},
  pages={1473--1487},
  year={2002},
  publisher={Elsevier}
}

@article{wellman2004theory,
  title={Theory of mind for learning and teaching: The nature and role of explanation},
  author={Wellman, Henry M and Lagattuta, Kristin H},
  journal={Cognitive development},
  volume={19},
  number={4},
  pages={479--497},
  year={2004},
  publisher={Elsevier}
}

@inproceedings{rabinowitz2018machine,
  title={Machine theory of mind},
  author={Rabinowitz, Neil and Perbet, Frank and Song, Francis and Zhang, Chiyuan and Eslami, SM Ali and Botvinick, Matthew},
  booktitle={International conference on machine learning},
  pages={4218--4227},
  year={2018},
  organization={PMLR}
}

@article{kosinski2024evaluating,
  title={Evaluating large language models in theory of mind tasks},
  author={Kosinski, Michal},
  journal={Proceedings of the National Academy of Sciences},
  volume={121},
  number={45},
  pages={e2405460121},
  year={2024},
  publisher={National Academy of Sciences}
}

@article{yang2026skillopt,
  title={Skillopt: Executive strategy for self-evolving agent skills},
  author={Yang, Yifan and Gong, Ziyang and Huang, Weiquan and Yang, Qihao and Zhou, Ziwei and Huang, Zisu and Li, Yan and Gao, Xuemei and Dai, Qi and Liu, Bei and others},
  journal={arXiv preprint arXiv:2605.23904},
  year={2026}
}

@article{alzubi2026evoskill,
  title={Evoskill: Automated skill discovery for multi-agent systems},
  author={Alzubi, Salaheddin and Provenzano, Noah and Bingham, Jaydon and Chen, Weiyuan and Vu, Tu},
  journal={arXiv preprint arXiv:2603.02766},
  year={2026}
}

@article{ni2026trace2skill,
  title={Trace2skill: Distill trajectory-local lessons into transferable agent skills},
  author={Ni, Jingwei and Liu, Yihao and Liu, Xinpeng and Sun, Yutao and Zhou, Mengyu and Cheng, Pengyu and Wang, Dexin and Zhao, Erchao and Jiang, Xiaoxi and Jiang, Guanjun},
  journal={arXiv preprint arXiv:2603.25158},
  year={2026}
}

@misc{deepseek2026v4,
      title={DeepSeek-V4: Towards Highly Efficient Million-Token Context Intelligence}, 
      author={DeepSeek-AI and Anyi Xu and Bangcai Lin and Bing Xue and Bingxuan Wang and Bingzheng Xu and Bochao Wu and Bowei Zhang and Chaofan Lin and Chen Dong and others},
      year={2026},
      eprint={2606.19348},
      archivePrefix={arXiv},
      primaryClass={cs.CL},
      url={https://arxiv.org/abs/2606.19348}, 
}

@misc{anthropic2026sonnet5,
  author       = {{Anthropic}},
  title        = {Introducing Claude Sonnet 5},
  year         = {2026},
  howpublished = {\url{https://www.anthropic.com/news/claude-sonnet-5}},
  note         = {Accessed: 2026-09-25}
}

@misc{google2026gemini38,
  author       = {Doshi, Tulsee and Popa, Raluca Ada},
  title        = {Introducing Gemini 3.8 Flash and 3.8 Flash Cyber},
  year         = {2026},
  howpublished = {\url{https://blog.google/innovation-and-ai/models-and-research/gemini-models/3-8-flash-and-3-8-flash-cyber/}},
  note         = {Google, accessed 2026-09-25}
}

@misc{openai2026gpt54mini,
  author       = {{OpenAI}},
  title        = {Introducing GPT-5.4 mini and nano},
  year         = {2026},
  howpublished = {\url{https://openai.com/index/introducing-gpt-5-4-mini-and-nano/}},
  note         = {Accessed: 2026-09-25}
}

@inproceedings{wang2025openhands,
  title={Openhands: An open platform for ai software developers as generalist agents},
  author={Wang, Xingyao and Li, Boxuan and Song, Yufan and Xu, Frank F and Tang, Xiangru and Zhuge, Mingchen and Pan, Jiayi and Song, Yueqi and Li, Bowen and Singh, Jaskirat and others},
  booktitle={International Conference on Learning Representations},
  volume={2025},
  pages={65882--65919},
  year={2025}
}

@article{belcak2025slm,
  title={Small language models are the future of agentic ai},
  author={Belcak, Peter and Heinrich, Greg and Diao, Shizhe and Fu, Yonggan and Dong, Xin and Muralidharan, Saurav and Lin, Yingyan Celine and Molchanov, Pavlo},
  journal={arXiv preprint arXiv:2506.02153},
  year={2025}
}

@misc{anthropic2026claudecode,
  author       = {{Anthropic}},
  title        = {Claude Code},
  howpublished = {\url{https://github.com/anthropics/claude-code}},
  note         = {GitHub repository, accessed 2026-09-25}
}

@misc{openai2026codex,
  author       = {{OpenAI}},
  title        = {Codex},
  howpublished = {\url{https://github.com/openai/codex}},
  note         = {GitHub repository, accessed 2026-09-25}
}

@misc{opencode2026,
  author       = {{OpenCode}},
  title        = {OpenCode: The Open Source AI Coding Agent},
  howpublished = {\url{https://opencode.ai/}},
  note         = {Accessed: 2026-09-25}
}

@misc{zechner2026pi,
  author       = {{Earendil Works}},
  title        = {Pi: AI Agent Toolkit},
  howpublished = {\url{https://github.com/earendil-works/pi}},
  note         = {GitHub repository, accessed 2026-09-25}
}
\clearpage
\appendix
\FloatBarrier
\section{Additional Method Details}
\label{app:method}
\subsection{Learner model}
For each teaching round, the student attempts a batch using its current library. The aware profile tracks successes and failures per category. For category $c$, its Beta-Bernoulli summary begins with $\alpha_c=\beta_c=1$. Observed successes increase $\alpha_c$ and failures increase $\beta_c$. The profile also stores failure counts and short teacher notes. Because the library changes across rounds, these statistics describe the student under an evolving curriculum. They should be read as estimates of the taught student, which differs from the bare model.

The analytical learner model stores a textual mechanism summary and a bounded set of hypotheses. Each hypothesis includes a claim, a confidence, supporting evidence, and a status. Additional fields hold inferred or measured task boundaries. The teacher uses this state to target a recurring execution problem. For example, it may suspect that repeated stock queries exhaust the action budget before all updates are issued. Analytical updates are stored separately from the empirical profile, which lets us examine observed behavior and its proposed explanation separately.

\paragraph{How the learner model changes a teaching decision.}
The empirical profile includes category performance, observed failure modes, teacher notes, and probe outcomes. These components can guide teaching before a detailed explanation is available. A low inventory success rate can motivate a lesson on completing stock updates within the action budget. Repeated intermediate reads can motivate consolidated execution. Missing nested files can motivate recursive discovery. The same task can support several plausible lessons. A student that omits necessary checks may need an explicit completion test, while a student that repeatedly checks finished work may need fewer checks. The learner model tells the teacher which of these lessons fits.

\subsection{Controlled probes}
Probes vary a supported task property while preserving the task family. Implemented axes include inventory size and file count for search tasks. The generator rebuilds each variant and recomputes its verifier. The bare student attempts a small ordered set of property values. Analytical hypotheses can guide these diagnostics, and empirical failures can trigger probes before any explanation exists. A transition from observed success to observed failure is recorded as a sampled boundary. For example, passing at 12 items and failing at 16 brackets a transition over those tested instances. All-pass, all-fail, and nonmonotonic sweeps are recorded as inconclusive.

\subsection{Skill generation and representation}
The teacher proposes up to three skills per round. The prompt requests concise reusable procedures and prohibits copying task-specific answers. Each body is limited to 900 characters. Skills have four native types: content, process, affordance, and plan. A record stores its body and target category together with harness scope, provenance, and a validation summary. Applicability metadata can name task families and failure modes, and an optional field holds a student-specific rationale.

The aware teacher receives the statistical profile, the available analytical learner model, and a focus list of weak categories. The agnostic teacher receives tasks, verifier information, eligible execution traces, and prior skills from its own condition. A generation request includes up to six task descriptions and four failed traces. Plan records can contain structured directives. In all complete packages reported here the delivery mode is advice, so the harness injects skill text and the student chooses its actions. The implementation also supports controller-enforced plans, which we leave for future evaluation.

\subsection{Empirical skill admission}
A candidate skill $s$ is evaluated on up to four development tasks matching its category. Each task is executed with the candidate and with an empty library, giving the standalone utility in Equation~\ref{eq:utility}. Candidate validation uses temperature zero. A candidate with no matched task is rejected, and a positive estimate admits the skill. Admission can deactivate a generic seed in the same category, and the library keeps at most two active skills per category. Up to two off-category tasks estimate interference with the same paired comparison. This quantity is logged for analysis only. Because the gate measures each candidate against an empty library, we evaluate complete libraries separately at test time.

\subsection{Selection after teaching}
The saved artifact contains the library and the learner model. The deployment experiment removes the teacher and evaluates three routes.
\begin{itemize}[leftmargin=*,itemsep=2pt]
\item \textbf{Student selection.} The student reads a metadata menu and returns at most two valid skill identifiers. It sees the task's category label, the first 400 characters of the instruction, and a short preview of each body. Selected bodies are injected before execution.
\item \textbf{Category routing.} A deterministic rule matches the task's category or suite tag and selects up to two skills by validation utility. Logs call this route \emph{oracle}.
\item \textbf{Flat injection.} Every active skill is supplied to the student.
\end{itemize}
Category routing uses task metadata and has no access to test outcomes. Deployment on untagged tasks is a natural extension.

\subsection{Evaluating the student model}
The teacher can predict bare-student success on calibration tasks. Brier score measures squared probability error. Predictions are scored on a panel of ten calibration tasks. Task success remains the primary measure of whether a curriculum helps the student.

\FloatBarrier
\section{Base Harness and Task Suites}
\label{app:tasks}
\paragraph{Base harness.}
The compact planning harness runs one episode per task. It first asks the student for a short plan, then enters a tool-calling loop of at most ten iterations. Each iteration lets the student call one tool or finish. Tool observations are truncated to 400 characters. The context keeps the system message and the six most recent conversational messages, so earlier observations drop out of view in long episodes. Supplied skills are injected as advisory text before the task instruction, and the student decides whether and how to follow them. The harness itself never changes during teaching. The same harness executes teaching, development, probe, and test tasks for every condition.

\paragraph{Task generation.}
Each suite has a seeded generator that builds a task instruction, its working environment, and a programmatic verifier. A frozen collection fixes the teaching, development, and test splits before any teaching begins, and all conditions within an experiment share one collection. Probes call the same generators with one property changed, such as the number of inventory items or the number of files to search, and rebuild the verifier for each variant. Table~\ref{tab:tasksuites} lists the suites and Table~\ref{tab:splits} gives split sizes.

\begin{table}[!htbp]
\centering\small
\caption{Task suites, categories, and what each verifier checks.}
\label{tab:tasksuites}
\begin{tabularx}{\linewidth}{l>{\raggedright\arraybackslash}X>{\raggedright\arraybackslash}X}
\toprule
Suite & Categories & Verifier checks \\
\midrule
Terminal & Word and line counting, extension renaming (with nested directories), sorted numeric merge, log deduplication, CSV filtering, staged file pipelines & Final files and directory state, or the reported count \\
Inventory API & Restocking every item below a threshold through list, query, and restock tools & Final stock levels and the completion marker \\
Arithmetic & Compound discounts, mixtures, relative rates & The final numeric answer \\
Python code & Function implementation and bug repair & The submitted function or repaired file \\
Dependent workflows & Chains of stages in which each stage reads the previous stage's output file & The output of every stage \\
\bottomrule
\end{tabularx}
\end{table}

\begin{table}[!htbp]
\centering\small
\caption{Split sizes per suite. Every test task is executed twice per condition.}
\label{tab:splits}
\begin{tabular}{lrrrl}
\toprule
Experiment & Teaching & Development & Test & Suites \\
\midrule
Main setting (S1 to S4) and controls & 12 & 6 & 12 & All five \\
Extended eight-round run & 40 & 12 & 30 & All five \\
Production harnesses & 12 & 6 & 12 & All except inventory API \\
\bottomrule
\end{tabular}
\end{table}

\paragraph{Generic initial skills.}
Agnostic and aware teaching start from the same six generic skills. The first three are \emph{count matching lines}, \emph{do stages in order}, and \emph{check then restock}. The other three are \emph{filter rows carefully}, \emph{merge and sort}, and \emph{rename matching files}. Learned skills can replace them within a category.

\FloatBarrier
\section{Curriculum Analysis and Paired Skill Artifacts}
\label{app:curriculum}
\subsection{Three views of a curriculum}
The S1 through S4 libraries contain 28 active learned skills, with 13 agnostic and 15 aware. Table~\ref{tab:skills} reports their native types. Aware teaching retains more process skills, and agnostic teaching uses more affordance skills. Complete bodies are available for ten agnostic and twelve aware skills from S2 through S4. S1 contributes metadata for its remaining six learned skills. The Qwen3 package adds one agnostic and four aware skills, which we use for category counts and paired cases. Native type records the teacher's representation choice, task category records where a skill is assigned, and functional labels record the behavior requested in the body.

\begin{table}[!htbp]
\centering\small
\caption{Native types of learned skills in S1 through S4. Generic initial skills are excluded.}
\label{tab:skills}
\begin{tabular}{lrrrrrr}
\toprule
Condition & Learned & Process & Content & Plan & Affordance & Complete bodies \\
\midrule
Agnostic & 13 & 6 & 3 & 1 & 3 & 10 \\
Aware & 15 & 10 & 3 & 2 & 0 & 12 \\
\bottomrule
\end{tabular}
\end{table}

Within S2 through S4, aware teaching deactivates seven of eighteen generic initial skills and agnostic teaching deactivates five. Each aware package retains two rename procedures.

\subsection{Functional annotation}
Table~\ref{tab:codebook} defines five functional labels. Coding uses the complete active body and permits overlapping labels. Annotations come from a single model-assisted pass and are recorded per skill. Counts describe this collection and treat skills within a library as related observations.

\begin{table}[!htbp]
\centering\small
\caption{Functional coding scheme. Native types and task categories remain separate fields.}
\label{tab:codebook}
\begin{tabularx}{\linewidth}{lX}
\toprule
Label & Inclusion rule \\
\midrule
Execution economy & Batch operations, reduce redundant tool calls, or allocate a limited action budget. \\
Environment coverage & Enumerate the complete target set or preserve coverage across directories. \\
State and progress & Track intermediate state or completed work to avoid stale or repeated actions. \\
Task semantics & Specify computation, exact transformations, or output-format requirements. \\
Verification and completion & Inspect a result or postcondition before proceeding or finishing. Conditional and deferred checks qualify. \\
\bottomrule
\end{tabularx}
\end{table}

\begin{table}[!htbp]
\centering\small
\caption{Functional labels in complete main-setting skill bodies (S2 through S4), as count (percent) within each condition. Labels overlap.}
\label{tab:functions}
\begin{tabular}{lrr}
\toprule
Function & Agnostic ($n=10$) & Aware ($n=12$) \\
\midrule
Execution economy & 5 (50.0) & 5 (41.7) \\
Environment coverage & 1 (10.0) & 6 (50.0) \\
State and progress & 3 (30.0) & 2 (16.7) \\
Task semantics & 6 (60.0) & 8 (66.7) \\
Verification and completion & 10 (100.0) & 10 (83.3) \\
\bottomrule
\end{tabular}
\end{table}

Within the complete bodies, the main-setting aware curriculum places more skills on environment coverage. Both conditions contain substantial advice about execution economy and verification. Native type can hide this overlap, since a plan and an affordance record both teach batching for dependent stages. Aware metadata names the compact planning harness as the skill's scope, while agnostic metadata typically declares general applicability.

\subsection{Allocation across tasks}
Table~\ref{tab:taskskills} shows the target categories of complete bodies. Six of twelve main-setting aware skills target renaming, and each aware library retains two of them. Four of ten agnostic skills target dependent workflows. All seven learned rename skills include environment coverage and verification. Both sorted-merge skills discuss semantics and verification.

\begin{table}[!htbp]
\centering\small
\caption{Active learned skills by target category. Main artifacts cover S2 through S4. The Qwen3 group uses a Qwen3-14B teacher and a Qwen3-8B student.}
\label{tab:taskskills}
\begin{tabular}{lrrrr}
\toprule
 & \multicolumn{2}{c}{Main artifacts} & \multicolumn{2}{c}{Qwen3} \\
\cmidrule(lr){2-3}\cmidrule(lr){4-5}
Category & Agnostic & Aware & Agnostic & Aware \\
\midrule
Extension rename & 1 & 6 & 0 & 0 \\
Dependent workflow & 4 & 2 & 0 & 1 \\
Inventory restock & 2 & 2 & 1 & 2 \\
Sorted merge & 1 & 1 & 0 & 0 \\
File pipeline & 1 & 1 & 0 & 0 \\
Mixture arithmetic & 1 & 0 & 0 & 0 \\
Compound discount & 0 & 0 & 0 & 1 \\
\midrule
Total & 10 & 12 & 1 & 4 \\
\bottomrule
\end{tabular}
\end{table}

\subsection{Paired teaching cases}
\label{sec:pairedskills}
We inspect paired cases through four kinds of evidence: the recorded student difficulty, the generated procedure, its validation, and final execution outcomes. Table~\ref{tab:pairedskills} summarizes them.

\begin{table}[!htbp]
\centering\small
\caption{Paired teaching cases. Acceptance percentages refer to the named task family within each run and measure the complete library.}
\label{tab:pairedskills}
\begin{tabularx}{\linewidth}{lXXrr}
\toprule
Case & Agnostic curriculum & Aware curriculum & Agnostic & Aware \\
\midrule
S4 inventory & Generic stock-check instruction remains active & Bulk stock snapshot replaces per-item discovery & 50.0 & 79.2 \\
Qwen3 inventory & Learned plan checks stock item by item & Two learned skills use bulk stock retrieval and explicit checks & 0.0 & 91.7 \\
S3 inventory & Learned plan skips routine verification & Learned procedure checks stock after all updates & 91.7 & 95.8 \\
S1 workflows & Skill for simple single-purpose commands & Skill for computing chain stages in one call & 75.0 & 95.8 \\
Extended renaming & Generic instruction with shallow glob & Recursive discovery with a final extension check & 20.0 & 100.0 \\
\bottomrule
\end{tabularx}
\end{table}

\paragraph{S4: a repair for a limited action budget.}
In S4, a first-round inventory teaching attempt reaches the ten-step limit before completion. The aware teacher then generates a bulk stock-check procedure. It receives standalone utility 0.50 on four matched development tasks and replaces the generic inventory skill. The agnostic library keeps the generic instruction to check stock and restock deficient items. The aware body specifies one stock snapshot followed by only the necessary updates, and it explicitly links redundant per-item queries to the student's action budget. The saved profile records difficulty with repeated verification and names consolidated stock checks as a useful procedure. Final inventory acceptance is 79.17\% for aware teaching, 50.00\% for agnostic teaching, and 62.50\% without skills. Execution records show per-item stock queries under the agnostic library and bulk queries under the aware skill.

\paragraph{Qwen3: a probe precedes the lesson.}
The Qwen3-8B student's first teaching batch includes two inventory step-limit failures and a missing completion marker. Before the first candidate is generated, the aware log records an inventory probe. The student passes at two items and fails at five or more. The teacher then generates a skill beginning with a full stock query. Its standalone validation succeeds on three of four matched tasks, while the bare comparison succeeds on none. The two retained aware inventory skills start with bulk retrieval and specify update verification. Their final inventory acceptance is 91.67\%, compared with 0\% for agnostic teaching and 8.33\% without skills. For one matched task, the agnostic record contains three SKU-list calls and seven per-item stock calls with no restock call. The aware record contains two bulk stock calls and one restock call. Both aware executions of that task pass, and both agnostic executions fail. This case gives the most complete chain in our artifacts, from observed failure through probing and validation to final success.

\paragraph{S3: choosing which verification to keep.}
S3 shows a subtler difference. Its initial inventory attempts include a step-limit failure and an understocked item. Both teachers eventually produce a bulk-read procedure. The agnostic body skips routine verification unless an update looks wrong. The aware body performs a final full-stock check and repairs remaining shortages before writing the completion marker. Both skills have standalone utility 0.25 on four matched tasks. Final inventory acceptance is 95.83\% with aware teaching and 91.67\% with agnostic teaching. The aware procedure keeps one well-placed check while still reducing repeated discovery calls.

\paragraph{S1: batching dependent stages.}
S1 retains an aware chain skill with standalone utility 0.50 that computes chain stages in one shell call. The agnostic library retains an affordance skill for simple single-purpose commands. Final dependent-workflow acceptance is 95.83\% for aware teaching and 75.00\% for agnostic teaching. Inventory acceptance is also higher under aware teaching, at 91.67\% against 62.50\%.

\paragraph{Useful lessons from agnostic teaching.}
Agnostic teaching also finds real procedural knowledge. An agnostic merge skill in S2 through S4 notices that the supplied merge inputs lack trailing newlines. Concatenating the files can join the last integer of one file to the first integer of the next, so the skill passes both files directly to the sorting command. Arithmetic in the main comparison also slightly favors agnostic teaching. The advantage of learner modeling is largest where failures depend on how the student acts through the harness.

\subsection{Verification as a conditional teaching decision}
Several chain skills in S2 through S4 combine reading, computation, and writing in one shell action. They discourage a separate read for every intermediate value and defer verification. The Qwen3 aware chain skill asks the student to check each output file before proceeding, and its inventory skills add explicit checks after updates. Within one main-setting aware library, the chain skill suppresses intermediate checks while the pipeline skill asks for one check per stage. These bodies expose a concrete teaching choice about which checks a student should perform under a constrained budget. A label such as verification is therefore most informative when paired with its timing and target.

\subsection{Admission support}
The aware rename procedures repeatedly address recursive discovery and exact extension replacement. Eight of twelve complete main-setting aware skills were admitted using one matched task, and the other four used four tasks. For agnostic teaching, three skills used one task, one used two, and six used four. Development splits contain one rename task each, which is reused across rename candidates. Larger development pools would sharpen utility estimates for narrow categories.

\FloatBarrier
\section{Teacher-Free Deployment Details}
\label{app:modes}
Table~\ref{tab:routeabs} gives absolute teacher-free results for the three DeepSeek libraries and the two local configurations. Each route executes its own no-skill baseline. L2 is the library from S4. None of these deployment scores enters Table~\ref{tab:main}.

\begin{table}[!htbp]
\centering\small
\caption{Frozen aware libraries evaluated with the teacher removed. Rates are percentages and differences are points. Each interval is the task-paired 95\% interval against the job's own baseline. Category is the route labeled oracle in the logs.}
\label{tab:routeabs}
\begin{tabular}{llrrl}
\toprule
Student and library & Route & Library & No skill & Difference and interval \\
\midrule
\Student{}, L1 & Student & 85.8 & 78.3 & $+7.5\;[-0.8,16.7]$ \\
 & Category & 90.0 & 80.8 & $+9.2\;[0.8,18.3]$ \\
 & Flat & 90.8 & 81.7 & $+9.2\;[2.5,16.7]$ \\
\addlinespace
\Student{}, L2 & Student & 78.3 & 79.2 & $-0.8\;[-8.3,6.7]$ \\
 & Category & 85.0 & 77.5 & $+7.5\;[0.8,15.0]$ \\
 & Flat & 84.2 & 73.3 & $+10.8\;[1.7,19.2]$ \\
\addlinespace
\Student{}, L3 & Student & 81.7 & 71.7 & $+10.0\;[1.7,19.2]$ \\
 & Category & 84.2 & 75.0 & $+9.2\;[0.8,18.3]$ \\
 & Flat & 86.7 & 73.3 & $+13.3\;[4.2,22.5]$ \\
\addlinespace
Qwen2.5-7B-Instruct & Student & 51.7 & 43.3 & $+8.3\;[0.0,16.7]$ \\
(Qwen2.5-32B teacher) & Category & 47.5 & 43.3 & $+4.2\;[-0.8,10.0]$ \\
 & Flat & 29.2 & 43.3 & $-14.2\;[-27.5,-0.8]$ \\
\addlinespace
Qwen3-8B & Student & 38.3 & 38.3 & $0.0\;[0.0,0.0]$ \\
(Qwen3-14B teacher) & Category & 53.3 & 38.3 & $+15.0\;[1.7,28.3]$ \\
 & Flat & 53.3 & 38.3 & $+15.0\;[0.0,30.0]$ \\
\bottomrule
\end{tabular}
\end{table}

Table~\ref{tab:extdeploy} reports teacher-free evaluation of the library from the extended run. All three routes improve on the 78.0\% no-skill baseline, and each interval excludes zero.

\begin{table}[!htbp]
\centering\small
\caption{Teacher-free deployment of the extended-run library. The no-skill baseline is 78.0\%.}
\label{tab:extdeploy}
\begin{tabular}{lrrr}
\toprule
Route & Acceptance & Gain & 95\% interval for gain \\
\midrule
Student selection & 83.7 & $+5.7$ & $[0.7,11.3]$ \\
Category routing & 85.7 & $+7.7$ & $[3.7,12.7]$ \\
Flat injection & 85.0 & $+7.0$ & $[2.7,11.7]$ \\
\bottomrule
\end{tabular}
\end{table}

\FloatBarrier
\section{Public Subsets and Self-Teaching}
\label{app:other}
The two experiments in this section are exploratory. Their caveats are stated beside each result.

\begin{table}[!htbp]
\centering
\caption{Public subsets and self-teaching. Acceptance in percent.}
\label{tab:public}
\setlength{\tabcolsep}{3pt}
\begin{tabular}{lrrr}
\toprule
Subset & None & Agn. & Aware \\
\midrule
MATH (3 runs) & 75.0 & 73.6 & \textbf{81.9} \\
GSM8K (2 runs) & 87.5 & 89.6 & \textbf{97.9} \\
\bottomrule
\end{tabular}

\vspace{6pt}
\begin{tabular}{lr}
\toprule
Self-teaching condition & Acc. \\
\midrule
No skill & 72.22 \\
Self-improve & 81.94 \\
Self-aware & 79.17 \\
External aware teacher & \textbf{83.33} \\
\bottomrule
\end{tabular}
\end{table}

\paragraph{Public subsets.}
The source lists contain 200 unique GSM8K questions and 500 unique MATH problems. The loader partitions each shuffled list into 50\% teaching, 20\% probe, and 30\% test pools. Each experiment uses twelve items from each pool, and every test item is executed twice. The answer verifier accepts a submission that contains the gold answer as a substring. This lenient rule applies equally to all conditions. Table~\ref{tab:publicruns} reports each run.

\begin{table}[!htbp]
\centering\small
\caption{Run-level acceptance on public subsets.}
\label{tab:publicruns}
\begin{tabular}{llrrr}
\toprule
Dataset & Run & No skill & Agnostic & Aware \\
\midrule
MATH & M1 & 79.2 & 70.8 & 83.3 \\
MATH & M2 & 70.8 & 79.2 & 83.3 \\
MATH & M3 & 75.0 & 70.8 & 79.2 \\
GSM8K & G1 & 91.7 & 83.3 & 100.0 \\
GSM8K & G2 & 83.3 & 95.8 & 95.8 \\
\bottomrule
\end{tabular}
\end{table}

Aware teaching matches or exceeds both baselines in all five runs.

\paragraph{Self-teaching.}
This experiment asks whether a strong external teacher adds value beyond cheaper alternatives that reuse the student's own backbone. We compare three teaching conditions against a no-skill baseline, all given identical task inputs and differing only in who produces the skills. In \emph{self-improvement}, the student reflects on its own failures and writes its own skills, with no separate teacher. In \emph{self-aware} teaching, a teacher reasons about the student but runs on the student's own backbone, a low-cost stand-in for a strong teacher. In \emph{aware} teaching, a strong external teacher model builds an explicit model of the student and generates skills from it. The setup has 36 task identities with two executions each, for 72 executions per condition. No-skill reaches 72.22\%. Self-improvement reaches 81.94\% (a gain of +9.7 points over no-skill, 95\% CI [+1.4, +19.4]). Self-aware teaching reaches 79.17\% (+6.9 points, 95\% CI $[-1.4, +18.1]$). The external aware teacher reaches 83.33\% (+11.1 points, 95\% CI [+1.4, +23.6]). Self-improvement and aware teaching each clear zero over the baseline; self-aware does not. Same-backbone self-improvement is therefore a strong baseline. The external teacher's margin over it is 1.39 points, which corresponds to one accepted execution out of 72, so on this single seed the advantage of the external teacher over self-improvement is small and should be read as exploratory. We also note that self-improvement incurred six skill-generation failures, whereas both teacher-driven conditions generated skills without errors.

\FloatBarrier
\section{Student-Model Diagnostics with API Students}
\label{app:modelstate}
GPT-5.4-Mini records accepted analytical updates in every teaching round. Table~\ref{tab:newcal} pairs its calibration scores with final task outcomes. Aware Brier is lower in two of three runs, and task accuracy is identical across the two teaching conditions within each run. For a strong student near ceiling, a better learner model improves prediction without changing the final score. This supports the teachability finding in Section~\ref{sec:rq3}.

\begin{table}[!htbp]
\centering\small
\caption{GPT-5.4-Mini comparisons with accepted analytical updates. Each calibration evaluation contains ten tasks. Task rates are percentages.}
\label{tab:newcal}
\begin{tabular}{rrrrrr}
\toprule
 & \multicolumn{3}{c}{Final task success} & \multicolumn{2}{c}{Final Brier} \\
\cmidrule(lr){2-4}\cmidrule(lr){5-6}
Run & No skill & Agnostic & Aware & Agnostic & Aware \\
\midrule
P1 & 90.8 & 88.3 & 88.3 & 0.077 & 0.051 \\
P2 & 89.2 & 90.0 & 90.0 & 0.137 & 0.055 \\
P3 & 86.7 & 93.3 & 93.3 & 0.065 & 0.066 \\
\midrule
Mean & 88.9 & 90.6 & 90.6 & & \\
\bottomrule
\end{tabular}
\end{table}

\FloatBarrier
\section{Extended Teaching and Skill Development}
\label{app:extended}
\paragraph{Configuration.}
This experiment pairs \Teacher{} with \Student{} in the compact planning harness with advice injection and matched retrieval. The frozen collection is generated at the hard difficulty setting with a fixed seed. It contains 40 teaching tasks, 12 development tasks, and 30 test tasks per suite across the same five suites. Eight rounds consume 200 distinct teaching tasks in batches of 25. Each condition executes every test task twice per round, so each rate averages 150 task scores.

\paragraph{Trajectory.}
Table~\ref{tab:ext-rounds} reports acceptance and calibration for every round. Each teaching policy proposes three skills per round, and the log records 24 generation attempts with no generation failure in either condition. Aware analytical updates are accepted in all eight rounds, and the final analytical learner model holds five hypotheses and a measured inventory boundary. The fixed calibration panel uses ten task identities disjoint from the development tasks reserved for skill validation.

\begin{table}[!htbp]
\centering\small
\caption{Extended teaching trajectory. Acceptance is a percentage over 150 tasks with two executions each. Brier and AUROC score the teacher's predictions of student success on the fixed ten-task panel.}
\label{tab:ext-rounds}
\setlength{\tabcolsep}{4pt}
\begin{tabular}{rrrrrrrr}
\toprule
 & \multicolumn{3}{c}{Test acceptance} & \multicolumn{2}{c}{Brier score} & \multicolumn{2}{c}{AUROC} \\
\cmidrule(lr){2-4}\cmidrule(lr){5-6}\cmidrule(lr){7-8}
Round & No skill & Agnostic & Aware & Agnostic & Aware & Agnostic & Aware \\
\midrule
1 & 74.3 & 78.3 & 78.3 & 0.175 & 0.132 & 0.938 & 0.938 \\
2 & 75.3 & 79.0 & 78.7 & 0.204 & 0.071 & 0.812 & 1.000 \\
3 & 78.0 & 80.0 & 80.7 & 0.181 & 0.046 & 0.938 & 1.000 \\
4 & 74.3 & 80.0 & 78.7 & 0.128 & 0.064 & 0.857 & 1.000 \\
5 & 76.7 & 79.0 & 80.3 & 0.111 & 0.063 & 0.844 & 1.000 \\
6 & 76.0 & 79.0 & 80.7 & 0.203 & 0.064 & 0.833 & 1.000 \\
7 & 77.3 & 80.3 & 82.3 & 0.108 & 0.079 & 0.812 & 1.000 \\
8 & 78.0 & 80.7 & 84.3 & 0.162 & 0.068 & 0.857 & 0.938 \\
\bottomrule
\end{tabular}
\end{table}

\paragraph{Paired intervals.}
The final aware advantage over no skill is 6.3 points with a paired bootstrap interval of $[2.0,11.0]$. The agnostic advantage over no skill is 2.7 points with an interval of $[-0.7,6.3]$. Only aware teaching separates from the untaught student. A direct comparison of aware with agnostic uses 100,000 paired task resamples, where each task keeps its two-execution mean under both conditions. It gives 3.7 points with a percentile interval of $[0.0,7.7]$. Thirteen tasks improve under aware teaching, six decline, and 131 are unchanged.

\paragraph{Where the gain arises.}
Table~\ref{tab:ext-suites} reports final acceptance by suite. Terminal tasks account for the advantage, contributing 4.7 points to the overall difference. Code is at ceiling under every condition, and mathematics and workflows are near ceiling.

\begin{table}[!htbp]
\centering\small
\caption{Final acceptance by suite in the extended run. Each suite contains 30 tasks with two executions each.}
\label{tab:ext-suites}
\begin{tabular}{lrrr}
\toprule
Suite & No skill & Agnostic & Aware \\
\midrule
Terminal & 45.0 & 50.0 & 73.3 \\
Inventory tool use & 56.7 & 56.7 & 55.0 \\
Mathematics & 91.7 & 96.7 & 95.0 \\
Code & 100.0 & 100.0 & 100.0 \\
Dependent workflows & 96.7 & 100.0 & 98.3 \\
\midrule
Overall & 78.0 & 80.7 & 84.3 \\
\bottomrule
\end{tabular}
\end{table}

\paragraph{Recursive renaming.}
A sixth-round teaching attempt fails because a nested file remains unrenamed. The aware teacher then generates a skill that enumerates matching files recursively and verifies that no original extension remains. Its admission test succeeds on a matching development task where the empty-library execution fails. On the test set, all five renaming tasks pass in both aware executions during each of the final three rounds. A representative final aware trace enumerates the \texttt{data/} tree with \texttt{find}, renames the discovered files, and checks for remaining matches. The corresponding agnostic trace uses a shallow glob and leaves a nested file unchanged.

\paragraph{Probe measurements enter skill content.}
A third-round inventory probe succeeds at eight stock-keeping units and fails at twelve and sixteen. A fourth-round skill explicitly cites the observed twelve-item failure. It advises consolidated stock retrieval and progress tracking, with final verification deferred when the budget is tight. This is a direct link from a probe measurement to lesson content. Agnostic teaching also produces advice about reducing stock queries in this run. Inventory acceptance ends at 55.0\% aware and 56.7\% agnostic, since the procedure still requires one update call per deficient item. A later probe succeeds at all tested sizes through sixteen, so a sampled boundary should be read as conditional on the tested instances.

\paragraph{Curriculum composition.}
The final agnostic library contains five learned and four generic skills. The aware library contains six learned and two generic skills, so aware teaching replaces more of its starting curriculum. Table~\ref{tab:ext-skills} lists every learned skill with its native type and standalone admission utility. All six aware skills are process skills, while agnostic teaching produces two process and three content skills. Aware skills cover renaming and pipelines in addition to the inventory, merge, and discount categories covered by both. Their mean utility is 0.58 against 0.35 for agnostic skills. Both teachers write a discount procedure and a budget-aware inventory procedure, so both conditions discover some useful lessons.

\begin{table}[!htbp]
\centering\small
\caption{Learned skills in the final extended libraries, paraphrased from their titles. Utility is the standalone admission estimate $\widehat U(s)$. Generic starting skills are excluded.}
\label{tab:ext-skills}
\begin{tabularx}{\linewidth}{llXlr}
\toprule
Condition & Round & Skill & Type & Utility \\
\midrule
Agnostic & 1 & Restock to the exact threshold & Content & $+0.25$ \\
Agnostic & 2 & Minimize API calls to fit the step budget & Process & $+0.25$ \\
Agnostic & 4 & Verify merge output line count and format & Process & $+0.50$ \\
Agnostic & 4 & Sanitize and cross-check merge inputs & Content & $+0.50$ \\
Agnostic & 7 & Compute sequential percentage discounts & Content & $+0.25$ \\
\midrule
Aware & 2 & Compute the exact restock delta per SKU & Process & $+0.25$ \\
Aware & 3 & Numeric merge with explicit line-count verification & Process & $+0.50$ \\
Aware & 3 & Verify each pipeline stage output before continuing & Process & $+1.00$ \\
Aware & 4 & Batch the restock loop to survive the step budget & Process & $+0.50$ \\
Aware & 6 & Recursive-safe file renaming with \texttt{find} & Process & $+1.00$ \\
Aware & 7 & Compute sequential percentage discounts & Process & $+0.25$ \\
\bottomrule
\end{tabularx}
\end{table}

\paragraph{Prediction quality.}
The aware Brier score is lower than the agnostic score in every round on the fixed panel. It reaches its minimum of 0.046 in round three and stays between 0.063 and 0.079 afterward. The aware AUROC is 1.0 in rounds two through seven, so in those rounds the aware teacher separates the panel tasks its student will pass from those it will fail without error. The final skill-validation Brier scores are similar at 0.250 aware and 0.253 agnostic. Those measurements use changing candidate-dependent tasks and answer a different question from the fixed panel.

\paragraph{Cost.}
The recorded teacher costs are \$1.245 for aware teaching and \$0.459 for agnostic teaching. Student costs are \$0.812 and \$0.739, and no-skill execution costs \$0.562. These totals cover teaching and repeated evaluation across all eight rounds.

\FloatBarrier
\section{Learner-Specificity Controls}
\label{app:controls}
\subsection{Component ablations}
Table~\ref{tab:ablation} reports the component ablations on run S2 with their own no-skill and agnostic baselines. Each ablated aware teacher still exceeds both baselines in its job.

\begin{table}[!htbp]
\centering\small
\caption{Component ablations on run S2. Acceptance in percent. Brier scores the teacher's predictions of student success on ten calibration tasks.}
\label{tab:ablation}
\begin{tabular}{lrrrr}
\toprule
Configuration & No skill & Agnostic & Aware & Aware Brier \\
\midrule
Full \method{} & 77.5 & 85.0 & 87.5 & 0.041 \\
No fine probes & 81.7 & 73.3 & 83.3 & 0.036 \\
No analytical learner model & 78.3 & 80.8 & 83.3 & 0.180 \\
\bottomrule
\end{tabular}
\end{table}

\subsection{False-profile control}
This control asks whether the teacher benefits from an accurate model of the specific student. The student is \Student{} and the teacher is \Teacher{}. The harness, frozen tasks, four teaching rounds, and admission procedure all stay fixed. Only the learner information supplied to the teacher changes. The overestimating profile describes the student as a frontier-class model comparable to GPT-5 or Claude-Opus-5. It claims near-ceiling competence across every task family, including tool use and long dependency chains. It also states that context loss and step limits are negligible, so repeated verification is cheap and most task families need little teaching. The underestimating profile gives the opposite description. It characterizes the student as comparable to a Qwen2-0.5B model with severe weaknesses in every skill from planning and arithmetic to state tracking. It tells the teacher that the learner needs highly explicit procedures, extensive decomposition, and detailed scaffolding. Both profiles stay frozen during teaching, so they remain wrong even as the teacher observes the real student's trajectories.

Table~\ref{tab:wrong-rounds} reports every round. Each run generates twelve candidates over four rounds without a generation failure, and none passes admission. All four final libraries therefore contain only the six generic seeds. Test acceptance still moves between rounds because each round re-executes the stochastic student. With a fixed generic library, final acceptance ranges from 71.7\% to 82.5\% across the four runs, which indicates the scale of evaluation noise at this sample size. The teacher costs per run are \$0.364, \$0.377, \$0.394, and \$0.408. The underestimating teacher's AUROC falls below 0.5 in most rounds, so its predictions invert the ordering of the student's real outcomes. The overestimating teacher ends with Brier between 0.19 and 0.29, several times the correct-profile value.

\begin{table}[!htbp]
\centering\small
\caption{False-profile trajectories. Acceptance (\%) on 60 test tasks with two executions each. Brier and AUROC score the teacher's predictions on the fixed ten-task panel.}
\label{tab:wrong-rounds}
\setlength{\tabcolsep}{2.5pt}
\begin{tabular}{llrrrrrrrrrrrr}
\toprule
 & & \multicolumn{4}{c}{Acceptance} & \multicolumn{4}{c}{Brier} & \multicolumn{4}{c}{AUROC} \\
\cmidrule(lr){3-6}\cmidrule(lr){7-10}\cmidrule(lr){11-14}
Profile & Seed & 1 & 2 & 3 & 4 & 1 & 2 & 3 & 4 & 1 & 2 & 3 & 4 \\
\midrule
Overest. & 6 & 79.2 & 76.7 & 78.3 & 71.7 & .288 & .284 & .380 & .190 & .50 & .83 & .79 & .69 \\
Overest. & 26 & 79.2 & 81.7 & 74.2 & 78.3 & .193 & .190 & .191 & .285 & .38 & .72 & .63 & .74 \\
Underest. & 6 & 78.3 & 78.3 & 76.7 & 82.5 & .644 & .558 & .646 & .746 & .50 & .29 & .41 & .13 \\
Underest. & 26 & 78.3 & 80.8 & 80.8 & 75.8 & .650 & .746 & .840 & .747 & .55 & .31 & .11 & .28 \\
\bottomrule
\end{tabular}
\end{table}

\subsection{Library swap between students}
This experiment asks whether a library fits the student it was written for. Two aware libraries come from the same teacher (\Teacher{}) and harness but different students, \Student{} and Gemini-3.8-Flash. At evaluation time we give each student its own library and the other student's library on the same held-out tasks. Each direction shares one no-skill baseline and reports all three routes. Category routing reduces retrieval quality as a confound, so it most directly measures whether library content fits its student. Table~\ref{tab:swap-full} reports both libraries against the shared baseline. The pattern is reciprocal: each student does better with its own library in all six comparisons.

\begin{table}[!htbp]
\centering\small
\caption{Library swap with gains over the shared no-skill baseline. Values are percentages and point differences with 95\% paired intervals.}
\label{tab:swap-full}
\setlength{\tabcolsep}{4pt}
\begin{tabular}{llll}
\toprule
Student and route & Own minus none & Swapped minus none & Own minus swapped \\
\midrule
\Student{}, student & $+0.0\;[-8.3,8.3]$ & $-4.2\;[-11.7,3.3]$ & $+4.2\;[-1.7,10.8]$ \\
\Student{}, category & $+14.2\;[7.5,21.7]$ & $+4.2\;[-4.2,13.3]$ & $+10.0\;[2.5,18.3]$ \\
\Student{}, flat & $+7.5\;[-2.5,17.5]$ & $+3.3\;[-5.0,12.5]$ & $+4.2\;[-4.2,13.3]$ \\
\addlinespace
Gemini-3.8-Flash, student & $+4.2\;[0.8,8.3]$ & $+0.8\;[-3.3,5.0]$ & $+3.3\;[0.8,6.7]$ \\
Gemini-3.8-Flash, category & $+5.8\;[0.8,10.8]$ & $+2.5\;[-1.7,6.7]$ & $+3.3\;[0.0,7.5]$ \\
Gemini-3.8-Flash, flat & $+5.8\;[0.8,10.8]$ & $+4.2\;[0.8,8.3]$ & $+1.7\;[0.0,5.0]$ \\
\bottomrule
\end{tabular}
\end{table}

\FloatBarrier
\section{Production Harnesses}
\label{app:harness}
\paragraph{Design.}
We extended the framework to run Claude Code, Codex, OpenCode, and Pi through their native command-line interfaces while preserving each harness's own execution workflow. All four harnesses use the same backbone through OpenRouter and receive the same frozen tasks, working-directory structure, execution budget, and evaluation procedure. We did not modify any internal planning loop, tool interface, or context-management policy. Changing them would turn the comparison into one between customized variants of each system. \method{} interacts with each harness only through controls that a normal user has. Learned skills are supplied as natural-language context together with the task instruction. The harness remains responsible for interpreting the skill, choosing its native tools, and producing the output. The evaluation covers four suites (terminal, arithmetic, code, and dependent workflows) with twelve test tasks each and two executions per task. Teaching runs four rounds as in the main setting.

\paragraph{Results.}
The harnesses differ from the base harness and from each other, so we report every result as a paired change against the same harness's no-skill run in the same round. Table~\ref{tab:harness} gives the final-round changes with intervals, Table~\ref{tab:harness-rounds} gives these changes after every round, and Table~\ref{tab:harness-suites} gives them by suite for the final round. Aware teaching ends at or above no skill in every harness. The largest risk appears in first-round agnostic libraries, which cost Pi 22.9 points and Codex 11.5 points before later rounds recover.

\begin{table}[!htbp]
\centering\small
\caption{Production harnesses with the \Student{} backbone. Final-round change (points) against each harness's own no-skill run, with paired 95\% intervals.}
\label{tab:harness}
\begin{tabular}{lll}
\toprule
Harness & Agnostic & Aware \\
\midrule
Codex & $+1.0\;[-5.2,7.3]$ & $+2.1\;[0.0,6.2]$ \\
Pi & $+1.0\;[-3.1,6.2]$ & $+2.1\;[0.0,6.2]$ \\
OpenCode & $0.0\;[-3.1,3.1]$ & $+1.0\;[0.0,3.1]$ \\
Claude Code & $-11.5\;[-22.9,-1.0]$ & $+5.2\;[-7.3,18.8]$ \\
\bottomrule
\end{tabular}
\end{table}

\begin{table}[!htbp]
\centering\small
\caption{Production harnesses: change in acceptance (points) against the same harness's no-skill run in each round. Each round evaluates 48 test tasks twice.}
\label{tab:harness-rounds}
\begin{tabular}{llrrrr}
\toprule
Harness & Condition & Round 1 & Round 2 & Round 3 & Round 4 \\
\midrule
Codex & Agnostic & $-11.5$ & $+3.1$ & $-1.0$ & $+1.0$ \\
 & Aware & $-2.1$ & $+4.2$ & $0.0$ & $+2.1$ \\
\addlinespace
Pi & Agnostic & $-22.9$ & $-1.0$ & $0.0$ & $+1.0$ \\
 & Aware & $+1.0$ & $0.0$ & $0.0$ & $+2.1$ \\
\addlinespace
OpenCode & Agnostic & $0.0$ & $+2.1$ & $+3.1$ & $0.0$ \\
 & Aware & $-1.0$ & $+1.0$ & $+3.1$ & $+1.0$ \\
\addlinespace
Claude Code & Agnostic & $+17.7$ & $+6.2$ & $+3.1$ & $-11.5$ \\
 & Aware & $+7.3$ & $+14.6$ & $-7.3$ & $+5.2$ \\
\bottomrule
\end{tabular}
\end{table}

\begin{table}[!htbp]
\centering\small
\caption{Production harnesses: final-round change (points) by suite against the same harness's no-skill run. Each suite has twelve tasks with two executions each.}
\label{tab:harness-suites}
\begin{tabular}{llrrrr}
\toprule
Harness & Condition & Terminal & Arithmetic & Code & Workflows \\
\midrule
Codex & Agnostic & $+4.2$ & $0.0$ & $0.0$ & $0.0$ \\
 & Aware & $+8.3$ & $0.0$ & $0.0$ & $0.0$ \\
\addlinespace
Pi & Agnostic & $+8.3$ & $-4.2$ & $0.0$ & $0.0$ \\
 & Aware & $+8.3$ & $0.0$ & $0.0$ & $0.0$ \\
\addlinespace
OpenCode & Agnostic & $0.0$ & $0.0$ & $0.0$ & $0.0$ \\
 & Aware & $0.0$ & $+4.2$ & $0.0$ & $0.0$ \\
\addlinespace
Claude Code & Agnostic & $-29.2$ & $-8.3$ & $+8.3$ & $-16.7$ \\
 & Aware & $-12.5$ & $+4.2$ & $0.0$ & $+29.2$ \\
\bottomrule
\end{tabular}
\end{table}

\FloatBarrier
\section{Implementation and Measurement Details}
\label{app:details}
\paragraph{Initialization and candidate lifecycle.}
The six generic initial skills cover line counting, staged pipelines, inventory updates, and three other terminal categories. Learned skills with positive standalone utility can supersede a generic skill in the same category. The category cap sorts by utility and keeps two active skills. In S2, a third rename candidate has utility 1.0 and stays inactive under the two-skill category cap.

\paragraph{Execution summaries.}
Result packages contain averaged final scores and compact execution summaries with tool counts. We use averaged task scores for all accuracy results and use tool-count summaries to illustrate behavior.

\paragraph{Budget accounting.}
Table~\ref{tab:budget} reports mean role totals for S2 through S4 at the configured prices. Student totals include preparation, development, and repeated test evaluation. The aware teacher spends \$0.39 per run, \$0.16 more than the agnostic teacher. The untaught student produces more than twice the output tokens of either taught student, and the aware student produces 56\% fewer output tokens even though its total also covers teaching and development executions. The complete aware condition costs \$0.67 per run. Local-model accounting omits accelerator time.

\begin{table}[!htbp]
\centering\small
\caption{Mean recorded role totals per run in S2 through S4.}
\label{tab:budget}
\begin{tabular}{llrrrr}
\toprule
Condition & Role & Calls & Input tokens & Output tokens & Estimated USD \\
\midrule
No skill & Student & 3725.33 & 1361517.3 & 1644371.3 & 0.443 \\
Agnostic & Student & 4185.00 & 1794873.3 & 805769.7 & 0.285 \\
Agnostic & Teacher & 76.00 & 54743.7 & 12194.0 & 0.231 \\
Aware & Student & 4355.33 & 1885968.7 & 724544.7 & 0.273 \\
Aware & Teacher & 82.67 & 119400.3 & 15304.0 & 0.392 \\
\bottomrule
\end{tabular}
\end{table}

\paragraph{Category outcomes.}
Table~\ref{tab:categories} joins S2 through S4 to their task categories. Category means weight task records equally. The largest aware gains over agnostic teaching appear in extension renaming and function implementation.

\begin{table}[!htbp]
\centering\small
\caption{Acceptance by task category in S2 through S4.}
\label{tab:categories}
\begin{tabular}{lrrrr}
\toprule
Category & Records & No skill & Agnostic & Aware \\
\midrule
Exact-word counting & 6 & 0.00 & 0.00 & 0.00 \\
Extension rename & 6 & 0.00 & 33.33 & 100.00 \\
Sorted merge & 6 & 50.00 & 75.00 & 75.00 \\
Log deduplication & 6 & 91.67 & 100.00 & 100.00 \\
CSV filtering & 6 & 100.00 & 100.00 & 100.00 \\
File pipeline & 6 & 50.00 & 66.67 & 50.00 \\
Inventory restock & 36 & 73.61 & 80.56 & 84.72 \\
Compound discount & 15 & 70.00 & 76.67 & 83.33 \\
Mixture arithmetic & 14 & 82.14 & 92.86 & 82.14 \\
Relative rate & 7 & 100.00 & 92.86 & 100.00 \\
Function implementation & 11 & 77.27 & 68.18 & 90.91 \\
Function tasks & 13 & 92.31 & 88.46 & 100.00 \\
Bug repair & 12 & 100.00 & 100.00 & 100.00 \\
Dependent workflow & 36 & 75.00 & 98.61 & 94.44 \\
\bottomrule
\end{tabular}
\end{table}

\paragraph{Preparation data.}
Each main run executes 60 distinct teaching records. Generation prompts expose subsets of task descriptions and trace summaries, and profiles and development feedback add further evidence. The 30 development records cover categories unevenly. Each run has one development task per terminal category and six inventory tasks, while code and arithmetic counts vary.

\paragraph{Evaluation protocol.}
Every teaching round evaluates the same frozen test split, and unless otherwise stated, results report the final round. This fixes the reporting point in advance and avoids selecting the best intermediate score. A task's score is the fraction of its two executions that the verifier accepts, so it is 0, 0.5, or 1. Acceptance is the mean task score over the held-out tasks, reported as a percentage, and run-level standard deviations are computed across runs.

\FloatBarrier
\section{Amortizing Preparation Cost}
\label{app:amortization}
For $N$ later tasks, the average model cost is
\begin{equation}
\overline C(N)=\frac{C_{\mathrm{prepare}}}{N}+C_{\mathrm{select}}+C_{\mathrm{execute}}.
\label{eq:cost}
\end{equation}
If a directly deployed stronger agent costs $C_{\mathrm{strong}}$ per task, teaching pays for itself once
\begin{equation}
N>\frac{C_{\mathrm{prepare}}}
{C_{\mathrm{strong}}-C_{\mathrm{select}}-C_{\mathrm{execute}}},
\end{equation}
given a positive denominator and acceptable accuracy. In the main setting, the aware teacher costs under one dollar per run. A deployment study under a common workload would measure both terms directly.

\end{document}